\documentclass[lettersize,journal]{IEEEtran}
\usepackage{amsmath,amsfonts}
\usepackage{algorithm}
\usepackage{algorithmic}
\usepackage{array}
\usepackage[caption=false,font=normalsize,labelfont=sf,textfont=sf]{subfig}
\usepackage{textcomp}
\usepackage{stfloats}
\usepackage{url}
\usepackage{verbatim}
\usepackage{graphicx}
\usepackage{cite}
\usepackage{booktabs}
\usepackage{soul}
\usepackage{color, xcolor}
\usepackage{hyperref}

\begin{document}

\title{E$^3$-UAV: An Edge-based Energy-Efficient Object Detection System for Unmanned Aerial Vehicles}




\author{Jiashun~Suo,~Xingzhou~Zhang,~Weisong~Shi,~\IEEEmembership{Fellow,~IEEE},~Wei~Zhou,~\IEEEmembership{Member,~IEEE}
\thanks{This work was supported in part by the National Natural Science Foundation of China under Grant 62162067, 62101480, and 62072434; in part by the Yunnan Province Science Foundation under Grant No.202005AC160007, No.202001BB050076; in part by the China Postdoctoral Science Foundation under Grant No.2021M693227; in part by the Innovation Funding of ICT, CAS under Grant No.E361040; and in part by the Beijing Natural Science Foundation under Grant No.4212027. \textit{(Corresponding author: Wei Zhou.)}}
\thanks{Jiashun Suo and Wei Zhou are with the Engineering Research Center of Cyberspace and the School of Software, Yunnan University, Kunming 650091, China (E-mail: suojiashun@mail.ynu.edu.com; zwei@ynu.edu.cn).}
\thanks{Xingzhou Zhang is with the Research Center of Distributed Systems, Institute of Computing Technology, Chinese Academy of Sciences, Beijing 100190, China (E-mail: zhangxingzhou@ict.ac.cn).}
\thanks{Weisong Shi is with the Department of Computer and Information Sciences, University of Delaware, Newark, DE 19716, USA (E-mail: weisong@udel.edu).}

\thanks{Copyright (c) 20xx IEEE. Personal use of this material is permitted. However, permission to use this material for any other purposes must be obtained from the IEEE by sending a request to pubs-permissions@ieee.org.}
\thanks{DOI: \href{https://doi.org/10.1109/JIOT.2023.3301623}{10.1109/JIOT.2023.3301623}}
}



\maketitle

\begin{abstract}
Motivated by the advances in deep learning techniques, the application of Unmanned Aerial Vehicle (UAV)-based object detection has proliferated across a range of fields, including vehicle counting, fire detection, and city monitoring. 
While most existing research studies only a subset of the challenges inherent to UAV-based object detection, there are few studies that balance various aspects to design a practical system for energy consumption reduction.

In response, we present the E$^3$-UAV, an edge-based energy-efficient object detection system for UAVs.
The system is designed to dynamically support various UAV devices, edge devices, and detection algorithms, with the aim of minimizing energy consumption by deciding the most energy-efficient flight parameters (including flight altitude, flight speed, detection algorithm, and sampling rate) required to fulfill the detection requirements of the task.
We first present an effective evaluation metric for actual tasks and construct a transparent energy consumption model based on hundreds of actual flight data to formalize the relationship between energy consumption and flight parameters.
Then we present a lightweight energy-efficient priority decision algorithm based on a large quantity of actual flight data to assist the system in deciding flight parameters.
Finally, we evaluate the performance of the system, and our experimental results demonstrate that it can significantly decrease energy consumption in real-world scenarios. Additionally, we provide four insights that can assist researchers and engineers in their efforts to study UAV-based object detection further.

\end{abstract}


\section{Introduction}
The utilization of Unmanned Aerial Vehicles (UAVs) for object detection has been widely implemented in various applications such as vehicle count~\cite{vehicle_count}, fire detection~\cite{fire_detection}, and city monitoring~\cite{city_monitoring}.
The advancement of artificial intelligence (AI) and edge computing~\cite{shi2016edge} has significantly motivated the development of UAV-based applications, which demonstrate extensive potential.
In particular, the deployment of edge devices on UAVs offers significant advantages in performing object detection tasks. Notably, these UAVs can operate in remote areas without internet access, thereby ensuring practicality.
Moreover, the utilization of edge devices enables the detection of unique facial features without transmitting data to cloud servers, which guarantees privacy~\cite{gao2021autonomous}.

Numerous researchers and engineers have dedicated their efforts to studying UAV-based object detection from various perspectives.
To aid researchers and engineers in addressing the challenges of UAV-based object detection, a large number of UAV-based datasets, including~\cite{zhu2018vision, standford, uav123, CARPK, au-air, asl-tid, BIRDSAI, FLAME}, have been introduced.
In addition, several detection algorithms, such as~\cite{algo_pre_1_rrnet, algo_pre_2_Towards, algo_pre_3_roi, algo_pre_4_TPH-YOLOv5, algo_pre_5_Uav-yolo, algo_pre_6_SyNet, algo_pre_7_zhang2019dense}, have been presented to enhance the detection accuracy of aerial view objects.
To optimize the energy consumption of flight or computation in UAV-based object detection tasks, several papers, such as~\cite{algo_energy_1_xu2019dac, algo_energy_2_deng2019energy, algo_energy_3_zhang2019skynet, communication_1_zeng2019energy, communication_2_yang2019energy, communication_3_zhou2018mobile, communication_4_li2020energy, communication_5_tran2020coarse}, have been published.
Furthermore, several works, including~\cite{sys_1_tijtgat2017embedded, sys_2_boubin2019managing, sys_3_li2019intelligent, sys_5_madasamy2021osddy, sys_6_nguyen2021visual, sys_7_lee2017real}, have developed systems for using UAVs to detect objects in the real world.
Despite these efforts, UAV-based object detection with edges still faces numerous challenges:

\IEEEpubidadjcol

\begin{itemize}
    \item \textbf{Limited detection precision.} The highest AP50 value of object detection achieved in the VisDrone-2019 challenge~\cite{visdrone2019} is 55.82\% using RRNet~\cite{algo_pre_1_rrnet}.
    Meanwhile, in the COCO test-dev~\cite{lin2014microsoft} dataset, the best performance is $80.8\%$ AP50, achieved using DINO~\cite{zhang2022dino}.
    Notably, VisDrone is a widely used UAV-based dataset, while COCO is a natural dataset.
    These detection results demonstrate that UAV-based object detection still faces significant challenges in developing effective detection algorithms.
    \item \textbf{Limited computing power of edge devices.} The detection speed of YOLOv3 with $608 \times 608$ input resolution is evaluated as 6.38 Frames Per Second (FPS) on the NVIDIA Jetson Xavier NX and 2.03 FPS on the NVIDIA Jetson Nano when using TensorRT for inference.
    Given the limited detection speed, UAV-based object detection may miss some objects during detection tasks.
    Therefore, the low detection speed is a significant challenge in UAV-based object detection tasks that needs to be addressed.
    \item \textbf{Limited energy.} Small quad-rotor drones have limited battery energy, which typically enables them to fly for only about 30 minutes.
    For example, the DJI Mavic Air 2 has a maximum flight time of 34 minutes, while the DJI Matrice M210 V2 can fly for up to 24 minutes with a takeoff weight of 6.14kg.
    If UAVs carry edge devices for object detection, the flight time will be further reduced due to the energy consumption of the edge devices.
    Therefore, finding ways to perform detection tasks with limited energy is a significant challenge in UAV-based object detection.
\end{itemize}

Unlike many other studies that only investigate specific limitations, we have developed the E$^3$-UAV system to balance the aforementioned limitations and optimize the performance of UAV-based object detection tasks.
This system is designed to be compatible with various UAVs, edges, and object detection algorithms.
The performance of UAV flight energy and edge-based object detection can be affected by factors such as flight altitude, flight speed, detection algorithm, and sampling rate.
Therefore, the primary objective of the E$^3$-UAV system is to strike a balance between these parameters and minimize energy consumption while fulfilling the task requirements.

We first developed a performance model that includes task completion profiling and system energy consumption profiling.
The task completion profiling addresses the issue of the limited ability of the mAP to evaluate the performance of UAV-based object detection in actual tasks. 
The system energy consumption profiling formalizes the relationship between energy consumption and various parameters such as UAV, edge device, flight altitude, flight speed, and detection sampling rate, thus guiding energy efficiency optimization.
We then proposed an energy efficiency-oriented decision algorithm, which uses conclusions derived from hundreds of flights to determine the optimal flight parameters.
This algorithm significantly reduces the search space, promoting search efficiency, and has a fast running speed and low energy consumption on edges.
We also designed a sampling rate search method and analyzed system parameters to aid users in employing the system.

We validated the effectiveness of the system through two parts.
First, we compared the energy consumption of the recommended flight parameter and other adjacent flight parameters.
Results showed that the recommended flight parameter reduced energy consumption by up to $17.07\%$ for car detection and $26.65\%$ for person detection compared to other adjacent flight parameters while meeting detection requirements.
Second, we used the best task completion score from several random flights as the task requirement to compare the energy consumption of the recommended flight parameter and random flights.
Results showed that the task completion score of the recommended flight parameter increased by $7.73\%$ with nearly the same energy consumption as the compared flight.
Furthermore, we introduced four insights from our experiments to aid researchers and engineers in further studying UAV-based object detection.

The main contributions in our work are as follows:
\begin{itemize}
    \item 
    We introduce the E$^3$-UAV, a novel UAV-based object detection system that enables the efficient completion of tasks through the selection of appropriate flight parameters.
    The system boasts a clear and structured design, ensuring compatibility with various UAVs, edge devices, and detection algorithms.
    Our experiments demonstrate that the E$^3$-UAV is capable of significantly reducing energy consumption while maintaining broad applicability.
    \item
    We present a transparent energy consumption model that explicates the intricate association between energy consumption and crucial factors, including UAV, edge device, flight altitude, flight speed, and detection sampling rate.
    Our model provides a comprehensive and methodical framework, which facilitates the optimization of energy consumption in UAV-based object detection.
    Furthermore, we augment the evaluation metric for detection tasks, which permits a more precise and pragmatic evaluation of UAV-based object detection performance in real-world scenarios.
    \item
    We propose an energy efficiency-oriented decision algorithm that facilitates the selection of flight parameters from a large pool of potential parameters.
    This algorithm effectively reduces the search space, improving the speed of decision-making.
    Additionally, the algorithm operates with very low energy consumption on edge devices.
    \item
    We provide four valuable insights regarding energy consumption, detection performance, and optimization strategies for UAV-based object detection.
    These insights are intended to enhance the understanding of researchers and engineers with respect to the practical implementation of UAV-based object detection in real-world scenarios.
\end{itemize}

The paper is structured as follows.
In Section II, we provide an overview of related work and the motivations for building our system.
In Section III, we present the performance modeling approach used in our work.
Specifically, we describe the benchmark dataset and the chosen object detection algorithm, followed by the definition of the task completion metric used in real-world scenarios.
Additionally, we propose an energy consumption profiling framework for UAVs, edge devices, and the entire system.
In Section IV, we describe the E$^3$-UAV system, including its structure and the energy efficiency-oriented decision algorithm. 
Section V presents the results of over a hundred flights used to fit the system parameters, validate the energy consumption model, and evaluate system performance.
Section VI discusses the conclusions and insights obtained from our experiments to assist researchers and engineers in further developing UAV-based object detection.
Finally, in Section VII, we conclude the work.

\section{Related work and motivation}

\subsection{UAV-based object detection algorithms}

Motivated by the rapid progress of deep learning, object detection has witnessed significant advancements.
Various novel object detection algorithms, such as R-CNN~\cite{rcnn}, Fast R-CNN~\cite{fastrcnn}, Faster R-CNN~\cite{fasterrcnn}, YOLO series~\cite{yolov1, yolov2, yolov3, yolov4}, and SSD~\cite{ssd}, have achieved remarkable performance on commonly used datasets such as COCO~\cite{lin2014microsoft}.
Nonetheless, the objects in the images from UAV perspective are often too small to be detected by standard object detection algorithms, which struggle to produce satisfactory results.
Consequently, several object detection algorithms designed specifically for the images from UAV perspective have been introduced.
The RRNet~\cite{algo_pre_1_rrnet} method combined anchor-free detectors with a re-regression module to construct the detector, achieving the highest AP50, AR10, and AR100 on the ICCV VisDrone2019 Object Detection in Images Challenge.
Yu \textit{et al.}~\cite{algo_pre_2_Towards} proposed the Dual Sampler and Head Detection Network (DSHNet), which was the first work aimed at resolving the long-tail distribution in UAV-based images.
To detect oriented objects in aerial images, Jian \textit{et al.}~\cite{algo_pre_3_roi} proposed the RoI Transformer, which achieved state-of-the-art performances on the DOTA~\cite{DOTA} and HRSC2016~\cite{HRSC2016} datasets.
Addressing the issues of variable object scales and motion blur on densely packed objects in UAV-based object detection, Zhu \textit{et al.}~\cite{algo_pre_4_TPH-YOLOv5} proposed TPH-YOLOv5, which achieved the state-of-the-art (SOTA) performance on the VisDrone Challenge 2021.
Liu \textit{et al.}~\cite{algo_pre_5_Uav-yolo} introduced UAV-YOLO, which outperformed standard object detectors in the detection of small objects captured by UAVs.
Albaba \textit{et al.}~\cite{algo_pre_6_SyNet} proposed SyNet, which combines multi-stage and single-stage detectors to decrease the high false-negative rate of multi-stage detectors and improve the quality of the single-stage detector, achieving SOTA results on the MSCOCO and VisDrone datasets.
Moreover, Zhang \textit{et al.}\cite{algo_pre_7_zhang2019dense} proposed a novel processing pipeline based on a cascade network to detect dense and small objects in UAV vision, achieving an average precision of $22.61\%$ on the test-challenge set in VisDrone-DET 2019.

\subsection{Energy efficiency optimization of algorithms on edges}

UAV-based object detection is a critical application with a high sensitivity to energy consumption.
Several efforts have been made to optimize the energy efficiency of algorithms on the edge.
For instance, Xu \textit{et al.}~\cite{algo_energy_1_xu2019dac} introduced the DAC-SDC low-power object detection challenge to advance energy-efficient object detection in UAV-based applications.
Deng \textit{et al.}~\cite{algo_energy_2_deng2019energy} proposed an energy-efficient system for real-time UAV-based object detection on an embedded platform that attained a speed of 28.5 FPS and an energy efficiency of 2.7 FPS/W on the DAC-SDC low power object detection challenge dataset.
Moreover, the SkyNet~\cite{algo_energy_3_zhang2019skynet} is an exceptionally lightweight deep neural network (DNN) comprising only 12 convolutional layers, which won the first place in both the GPU and FPGA tracks of the 56th IEEE/ACM Design Automation Conference System Design Contest (DAC-SDC).

\subsection{Latency optimization of algorithms on edges}

The deployment speed of object detection algorithms is crucial due to the limited computing resources at the edge.
The DroNet~\cite{algo_deployment_1_kyrkou2018dronet} achieves a frame rate ranging from 5 to 18 FPS on various platforms for detecting vehicles in aerial images.
To detect vehicles in real-time, ShuffleDet~\cite{algo_deployment_2_majid2018shuffledet} uses inception modules and deformable modules with channel shuffling and grouped convolutions, running at a speed of 14 FPS on the NVIDIA Jetson TX2.
Balamuralidhar \textit{et al.}~\cite{algo_deployment_3_balamuralidhar2021multeye} proposed MultEYE, which detects, tracks, and estimates the velocity of vehicles in a sequence of aerial images, achieving 29 FPS on Nvidia Xavier NX with a 512 × 320 frame resolution.
Zhang \textit{et al.}~\cite{algo_deployment_4_zhang2019slimyolov3} presented SlimYOLOv3, an efficient deep object detector through channel pruning of convolutional layers.
SlimYOLOv3 is two times faster than YOLOv3 with a $90.8\%$ reduction in FLOPs and a $92.0\%$ decrease in parameter size, while maintaining comparable detection accuracy.

\subsection{UAV energy consumption optimization}

Numerous studies have focused on optimizing energy consumption through communication and flight trajectory optimization. 
For example, Zeng \textit{et al.}~\cite{communication_1_zeng2019energy} formulated an energy minimization problem by jointly optimizing the UAV trajectory and communication time allocation among ground nodes (GNs).
Yang \textit{et al.}~\cite{communication_2_yang2019energy} investigated a wireless communication system enabled by UAVs with energy harvesting to minimize energy consumption while fulfilling the minimal data transmission requirements of users.
From an energy efficiency perspective, Zhou \textit{et al.}~\cite{communication_3_zhou2018mobile} proposed a joint route planning and task assignment approach, significantly improving energy efficiency.
Li \textit{et al.}~\cite{communication_4_li2020energy} optimized computation offloading by simultaneously optimizing the UAV trajectory, user transmit power, and computation load allocation, resulting in maximum UAV energy efficiency. 
Finally, Tran \textit{et al.}~\cite{communication_5_tran2020coarse} utilized heuristic search and dynamic programming to design the UAV trajectory, meeting both the requested timeout requirement and energy budget while minimizing total energy consumption, and their results outperformed state-of-the-art benchmarks in terms of energy consumption and outage performance.

\subsection{UAV-based object detection system}

Various systems have been developed for UAV-based object detection tasks, as they have wide application value.
Tijtgat \textit{et al.}~\cite{sys_1_tijtgat2017embedded} evaluated the performance of YOLOv2 on an NVIDIA Jetson TX2 for a UAV-based warning system.
Boubin \textit{et al.}~\cite{sys_2_boubin2019managing} presented a model-driven approach to managing Fully autonomous aerial systems (FAAS) and released an open-source FAAS suite, SoftwarePilot. 
Li \textit{et al.}~\cite{sys_3_li2019intelligent} developed a UAV-based object detection system that deployed Tiny-YOLO on a mobile device to detect objects captured by a UAV, and Madasamy \textit{et al.}~\cite{sys_5_madasamy2021osddy} proposed deep YOLO V3, implemented on an embedded system, for object detection. 
Nguyen \textit{et al.}\cite{sys_6_nguyen2021visual} presented a real-time fire detection system comprising a low-cost camera, a lightweight companion computer, a flight controller, and localization and telemetry modules. 
Lee \textit{et al.}\cite{sys_7_lee2017real} utilized cloud computing to build a UAV-based object detection system capable of applying hundreds of object detection algorithms in near real-time, taking advantage of the high computing power of the cloud.

\subsection{Motivation}

\begin{table*}[ht]
\centering
\caption{The functions of various UAV-based object detection systems}
\scalebox{0.8}{
\begin{tabular}{c|ccccc}
\toprule
\textbf{System} & \textbf{Detection Model} & \textbf{Algorithm Deployment on Edges} & \textbf{Edge Energy Optimization} & \textbf{Flight Status Optimization} & \textbf{Energy Consumption Modeling} \\
\midrule
\midrule
Nils Tijtgat \textit{et al.}~\cite{sys_1_tijtgat2017embedded} & \checkmark & \checkmark &  &  &   \\
SoftwarePilot~\cite{sys_2_boubin2019managing} &   & \checkmark &   & \checkmark & \checkmark \\
Li Chuanlong \textit{et al.}~\cite{sys_3_li2019intelligent} & \checkmark & \checkmark &  &  & \\
OSDDY~\cite{sys_5_madasamy2021osddy} & \checkmark & \checkmark &  &  &  \\
A. Q. Nguyen \textit{et al.}~\cite{sys_6_nguyen2021visual} & \checkmark & \checkmark &  &  &  \\
Jangwon Lee \textit{et al.}~\cite{sys_7_lee2017real} & \checkmark &  &  &  &  \\
\textbf{E$^3$-UAV} & \checkmark & \checkmark & \checkmark & \checkmark & \checkmark \\
\bottomrule
\end{tabular}}
\label{table_UAV_system}
\end{table*}

Although numerous systems have been proposed for UAV-based object detection, there are still several areas that require improvement. 
\textbf{To conserve energy, a UAV-based object detection system that relies on edge computing must take into account several factors:}
(1) \textbf{Detection models.} 
The selection of detection models is critical as different models have varying detection accuracy and energy consumption levels.
Therefore, a suitable model should be chosen to decrease energy consumption while maintaining detection accuracy.
(2) \textbf{Deployment methods on edges.}
Different deployment methods, such as various deployment frameworks or quantification methods, can result in different energy consumption.
Therefore, a proper deployment method can decrease system energy usage.
(3) \textbf{Flight status.}
Different flight speeds and altitudes can result in different flight trajectories that consume different energy. 
Thus, flight status is a crucial consideration in energy conservation. 
(4) \textbf{Energy consumption modeling.}
An energy consumption model can provide guidance for optimizing system energy efficiency.
Creating an appropriate energy consumption model for detection tasks is critical.

However, while many existing systems take into account some of the aforementioned factors (as summarized in Table~\ref{table_UAV_system}), it remains a challenge to consider all of these factors simultaneously as they often interact with one another, making it difficult to achieve optimal energy consumption.
For instance, opting for a lighter detection model to reduce energy consumption may lead to a reduction in detection precision, which can be compensated by flying the UAV at a lower altitude or slower speed, but this would increase energy consumption while performing the same detection task, potentially resulting in an overall increase in energy consumption.
In this study, we aim to strike a balance between these various factors to minimize energy consumption.
To this end, we propose the E$^3$-UAV system for real-world detection tasks, which takes into account all relevant factors and optimizes energy consumption accordingly.
The system supports a range of detection algorithms, UAVs, and edge devices, making it highly practical for real-world applications.

\textbf{We hope our work will provide valuable insights to researchers and engineers on UAV-based object detection in various fields}, such as (1) the impact of flight altitude and speed on energy consumption, (2) the influence of algorithms and edge devices on energy consumption, (3) the performance of current representative object detection algorithms in actual UAV-based tasks, and (4) the comprehensive methods for energy efficiency optimization in UAV-based object detection.

\section{Performance modeling}
In this section, we introduce: (1) the selection of the object detection algorithms and associated dataset, (2) the definition of the task completion metric, and (3) the introduction of energy consumption profiling for UAVs, edge devices, and the overall system.

\subsection{Object detection algorithm and dataset}\label{sec3_1}
We have selected the representative object detection algorithm, YOLOv3, as the benchmark algorithm in this study.
The performance of YOLOv3 in UAV-based object detection tasks is influenced by various input resolutions.
Larger input resolutions enhance the detection capability for smaller objects, offering an advantage for UAVs operating at higher altitudes to identify objects.
Consequently, larger input resolutions enable UAVs to detect a broader area at increased flight altitudes, conserving detection energy per square kilometer.
However, larger input resolutions also raise the energy consumption of edge devices, as they necessitate more computational processing at higher power.
To minimize system energy consumption while accomplishing the task objective, a suitable input resolution must be selected through a trade-off between detection performance and energy consumption (see Section \ref{sec4_2} for details).

Deployment optimization techniques can enhance the performance of models on GPU-based edge devices.
Our previous work~\cite{suo2021feasibility} details the optimization of YOLOv3 deployment performance on edge devices.
In this work, the system employs the TensorRT inference engine and 16-bit quantification to improve deployment performance.
We have selected four different input resolutions ($416 \times 416$, $608 \times 608$, $800 \times 800$, and $1024 \times 1024$) as the detection models within the system.
It is important to note that higher resolutions yield greater precision, albeit at the cost of increased latency.

The VisDrone dataset~\cite{zhu2018vision} is utilized to train YOLOv3 with various input resolutions.
Comprising 10,209 static images captured by diverse drone-mounted cameras, the VisDrone dataset covers an extensive range of factors, such as location (spanning 14 different cities across China), environment (urban and rural), objects (pedestrians, vehicles, bicycles, etc.), and density (sparse and crowded scenes).
As a result, the VisDrone dataset can significantly enhance UAV-based detection performance in real-world scenarios.

We have modified the anchor box~\cite{yolov3} of YOLOv3 to optimize detection accuracy.
Initially, we observed that the anchor box computed by k-means in the VisDrone dataset was too small, providing insufficient space for object coordinate calculation.
To address this issue, we doubled the size of the anchor box, which our experiments indicate has considerably improved the mean Average Precision (mAP).
Furthermore, we have proportionally resized the anchor box in accordance with the changes in input resolution, ensuring that each model has the most suitable anchor box.
For instance, if an anchor box is (2, 5) with a $416 \times 416$ input resolution, the anchor box would be set to (4, 10) with an $832 \times 832$ input resolution.

\subsection{Task completion profiling}
The mean Average Precision (mAP) is a widely recognized metric in object detection for evaluating algorithm performance on datasets.
As defined in the COCO metric~\cite{lin2014microsoft}, mAP is calculated based on the integral of the precision-recall (P-R) curve, accounting for both precision and recall.
However, mAP has limitations in effectively profiling detection performance in real-world tasks.
Firstly, actual detection tasks are more akin to detecting objects in videos, while mAP is an image-based evaluation metric, rendering it unsuitable for evaluating algorithms in videos.
Secondly, in real-world tasks, recall is often prioritized.
For instance, in rescue tasks, failing to detect a person may result in dire consequences, whereas misidentifying an object, such as a stone as a person, does not directly impact the rescue mission.
In the context of mAP evaluation, if an algorithm detects half of the objects without error or detects all objects with a $50\%$ error rate (the order is one right and one error), both cases yield an mAP of $50\%$.
Such an equal score for these two scenarios is unreasonable in recall-centric tasks.

In this study, we define a new precision and recall metric to more accurately evaluate detection performance in real-world tasks.
The precision reflects the proportion of correctly detected objects by algorithms, calculated using all sampled images captured by a camera during tasks.
Equation \ref{formula_precision} outlines the calculation of precision, where $TP1$ denotes the number of correctly detected objects and $FP$ represents the number of incorrectly detected objects in all images.
A correct object must satisfy two conditions: (1) the object classification is accurate, and (2) the Intersection over Union (IoU) of the prediction box and label box exceeds 0.5.

\begin{equation}
Precision_{new} = \frac{TP1}{FP+TP1}
\label{formula_precision}
\end{equation}

The recall represents the proportion of objects detected among all objects in a task.
Consequently, the recall calculation is based on a video, as demonstrated in Equation \ref{formula_recall}.
The variable $TP2$ denotes the number of objects found in the video.
It is important to note that an object is considered found as long as it is correctly detected once; for example, an object is deemed found if it is accurately detected in the current frame but not in the subsequent frame.
The $FN$ refers to the number of objects not found in the video.

\begin{equation}
Recall_{new} = \frac{TP2}{FN+TP2}
\label{formula_recall}
\end{equation}

To distinguish from the conventional metric, the precision and recall defined by the system will be referred to as E$^3$-UAV precision and recall in the subsequent sections.

\subsection{UAV energy consumption profiling}\label{sec3_3}
\begin{figure}[ht]
\centering
\includegraphics[width=3.4in]{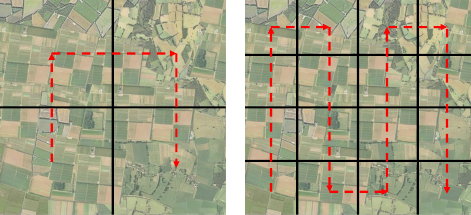}
\caption{The flight path of UAVs at varying altitudes is depicted by the red line, while the uniform square segments represent the detection zones as captured by the camera at higher altitude (left) and lower altitude (right). It can be noted that an increased altitude corresponds to an expanded field of view, consequently leading to a reduced detection path length.}
\label{flight_area}
\end{figure}

We denote the flight altitude as $h$, and $\theta_{x}$ and $\theta_{y}$ as the shooting angles of the camera in the $x$ and $y$ directions, respectively.
The detection area of the camera is represented by Formula \ref{formula_camera_area}.
At a constant camera angle, a lower flight altitude $h$ corresponds to a smaller detection area $S_{detection}$ according to Formula \ref{formula_camera_area}.
In Fig. \ref{flight_area}, assuming the total detection area is $S$ and the area of each small square is $S_{detection}$, a higher flight altitude covers a broader detection range, thus reducing the flight path required for detecting the entire area.
The longer flight path will consume more energy, thus increasing flight altitude can conserve energy consumption.

\begin{equation}
\begin{aligned}
S_{detection}&=2htan\theta _{x}\cdot 2htan\theta _{y}\\&=4tan\theta _{x}tan\theta _{y}h^{2}
\end{aligned}
\label{formula_camera_area}
\end{equation}

To formalize the UAV energy consumption $P_{UAV}$ with respect to flight altitude and speed, we define the unit of $P_{UAV}$ as $Wh/km^2$, which represents the energy consumption per square kilometer for UAV detection.
Our calculation proceeds as follows.
Firstly, we measure the horizontal length $l$ of the detection area at different altitudes.
Secondly, we measure the straight flight time $t$ at various speeds $v$ and altitudes with the same energy consumption $E$.
The detection area $S$ is equal to $lvt$.
Formula \ref{formula_uav_power} illustrates the calculation of $P_{UAV}$.
Given a constant camera, $l$ is solely dependent on the flight height $h$, allowing $l$ to be denoted by $\theta h$, where $\theta$ is a camera parameter determined experimentally.
According to our experiments, \textbf{the influence of speed and altitude on average flight time per watt-hour can be disregarded in actual flights} (proof is provided in Section \ref{sec6_1}).
This observation suggests that, within the civilian range (flight altitude below 120 meters), flight time is primarily determined by the inherent attributes of UAVs.
Consequently, when a UAV and camera are specified, $\theta t$ can be represented as a single parameter, $\alpha$, which denotes the UAV parameter. 
In our experiments, we employ a substantial amount of actual flight data to fit the $\alpha$ value.
Subsequently, we can compute the UAV energy consumption for various altitudes and speeds.

\begin{equation}
P_{UAV} = \frac{E}{S} = \frac{E}{lvt} = \frac{E}{\theta hvt} = \frac{E}{\alpha hv}
\label{formula_uav_power}
\end{equation}

\subsection{Edge device energy consumption profiling}
To facilitate energy modeling, we aim for the edge device energy consumption unit to be consistent with $P_{UAV}$, which is $Wh/km^2$.
We calculate the time required for detecting one kilometer (with the unit in seconds) at different altitudes and speeds, as demonstrated in Formula \ref{formula_edge_time}.
The system does not continuously execute detection algorithms; instead, it operates at varying sampling rates to satisfy the detection demands in detection tasks, conserving edge device operational energy consumption.
By measuring the average power of edge devices with algorithms, we convert the running power $P_{run}$ and standby power $P_{standby}$ into units of $Wh/s$.
We use $r$ to represent the sampling rate, indicating the frequency of running the algorithm per second, and $r_{max}$ to denote the maximum frequency of running the algorithm per second, which corresponds to the Frames Per Second (FPS).
The edge device energy consumption $P_{edge}$ comprises both running and standby energy consumption, as illustrated in Formula \ref{formula_edge_power}.

\begin{equation}
t = \frac{10^6}{lv} = \frac{10^6}{\theta hv}
\label{formula_edge_time}
\end{equation}

\begin{equation}
\begin{split}
P_{edge} &= \frac{r}{r_{max}}tP_{run} + (1 - \frac{r}{r_{max}})tP_{standby}\\
&= \frac{10^6r}{\theta hvr_{max}}P_{run} + (1 - \frac{r}{r_{max}})\frac{10^6}{\theta hv}P_{standby}
\end{split}
\label{formula_edge_power}
\end{equation}

\subsection{System energy consumption profiling}
The system energy consumption is the sum of UAV and edge device energy consumption, as depicted in Formula \ref{formula_system_power}, since they share the same unit.

\begin{equation}
P_{system} = P_{UAV} + P_{edge}
\label{formula_system_power}
\end{equation}

Two aspects contribute to the system energy consumption:

Firstly, \textbf{static factors}, which pertain to hardware selection, include:
\begin{itemize}
\item Camera properties: Different cameras possess distinct camera parameters, $\theta$, which influence the detection area, consequently affecting flight time and both UAV and edge device energy consumption.
\item UAV properties: Various UAVs have different UAV parameters, $\alpha$, which impact UAV energy consumption when camera properties are fixed.
\item Edge device properties: Diverse edge devices exhibit different standby power $P_{standby}$, running power $P_{run}$, and maximum sampling frequency $r_{max}$, affecting the edge device energy consumption.
\end{itemize}

Secondly, \textbf{dynamic factors}, which concern flight and detection parameter selection, include:
\begin{itemize}
\item UAV flight altitude: A higher flight altitude provides a broader detection area, thereby reducing energy consumption per square kilometer of detection.
\item UAV flight speed: A faster flight speed can also increase the detection area while maintaining the same energy consumption, consequently decreasing energy consumption per square kilometer of detection.
\item Object detection algorithm: Different object detection algorithms possess varying computational complexities, affecting $P_{run}$ and, in turn, the running energy consumption of edge devices.
\item Sampling rate $r$: Reducing the sampling rate can increase the proportion of standby time in a task, resulting in a longer standby duration and reduced running time. This leads to lower energy consumption by the edge device since $P_{run}$ is greater than $P_{standby}$.
\end{itemize}

\begin{figure}[ht]
\centering
\includegraphics[width=3.4in]{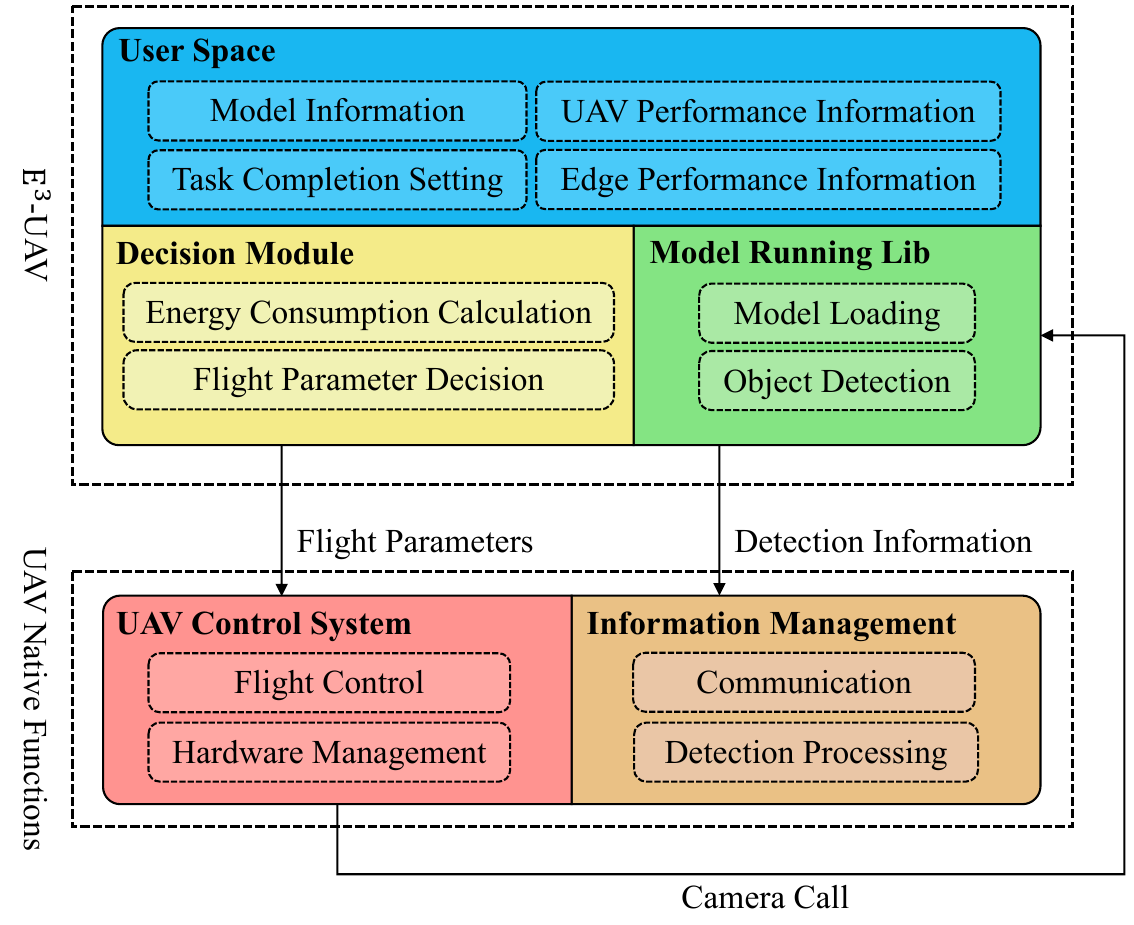}
\caption{The E$^3$-UAV system structure.}
\label{fig_system_structure}
\end{figure}

When hardware is predetermined, static parameters cannot be altered to conserve energy.
Consequently, the system aims to balance dynamic parameters to minimize energy consumption.
A higher flight altitude, faster flight speed, detection algorithms with lower computational complexity, and reduced sampling rate can decrease energy consumption.
However, these factors may also diminish detection performance, as a higher flight altitude renders objects too small to detect, a faster flight speed and lower sampling rate might cause UAVs to miss objects, and detection algorithms with lower computational complexity exhibit reduced detection accuracy.
Thus, the system requires an effective method for determining dynamic parameters by striking a balance between detection performance and energy consumption.

\section{E$^3$-UAV}

This section presents the E$^3$-UAV system, encompassing the system structure and decision method. 
The system's design principle requires support for various UAV platforms, edge devices, and detection algorithms.

\subsection{System structure}

Figure \ref{fig_system_structure} illustrates the E$^3$-UAV structure, which comprises three components: user space, decision module, and model running library.

User space facilitates system utilization with different configurations, incorporating four parts: model information, UAV performance information, edge performance information, and task completion settings.
The model information interface enables users to define object detection model specifications, including the model name, latency, running energy consumption on different edges, and E$^3$-UAV precision and recall with varying sampling rates and altitudes at a specific speed.
Users can load one or more detection models into the system using these configurations.
The UAV performance information interface permits users to define configurations for different UAVs equipped with cameras, including benchmark energy consumption $E$, camera parameter $\theta$, UAV parameter $\alpha$, and maximum flight speed.
The system supports various UAVs using the provided UAV performance information.
The edge performance information interfaces allow for defining the standby energy consumption of edge devices, enabling system compatibility with different edges.
Users can establish different task completion scores for various detection tasks using the task completion settings interface.

The main functions of the decision module include energy consumption calculation and flight parameter decision.
The energy consumption calculation function calculates the system energy consumption for different flight parameters based on user space configurations.
The flight parameter decision function decides on flight parameters that meet task completion requirements while prioritizing energy efficiency.
The decision method will be described in Section \ref{sec4_2}.
Flight parameters include flight altitude, flight speed, detection model, and sampling rate.
Subsequently, the model running lib and UAV control system operate based on the flight parameters.

The model running lib is responsible for loading and executing detection models.
We provide YOLOv3 deployment for TensorRT, TensorFlow, and PyTorch to support various edge devices.
Users merely need to supply the YOLOv3 weight file they have trained for deployment.
For other algorithms, users can provide their deployment file to deploy the models using our interface.
We plan to offer additional algorithms in the future.
The model loading function loads the detection model recommended by the decision module.
The object detection function calls the UAV camera to detect objects using the model loaded by the loading function with the sampling rate suggested by the decision module.

In a detection task, the user first establishes configurations, and then the decision module determines suitable flight parameters.
The UAV can automatically fly at an energy-efficient altitude and speed if it possesses an automatic flight system; otherwise, it can be manually flown at the recommended altitude and speed.
Simultaneously, the model running lib calls object detection models based on the flight parameters to obtain detection information.

The E$^3$-UAV provides flight parameters and detection information to UAV self-functions, such as the UAV control system and information management module.
The UAV control system can utilize the flight speed and altitude offered by the E$^3$-UAV to direct the UAV in performing the task, minimizing energy consumption.
Users can program the information management application to use detection information for custom requirements.
The communication function can transmit detection information to the task management center, which can aggregate information from multiple UAVs, enhancing the efficiency of detection tasks.
For specific tasks, such as traffic flow monitoring, users can employ a dedicated detection processing function to calculate real-time traffic flow using detection information on the UAV.

\subsection{Energy efficiency-oriented decision algorithm}\label{sec4_2}

The E$^3$-UAV determines flight parameters (flight altitude, flight speed, detection model, and sampling rate) with minimal energy consumption to fulfill the task requirements.
\textbf{This decision-making process involves four key points:}

(1) Flight parameters have four dimensions, which influence one another.
For instance, increasing flight altitude leads to reduced UAV energy consumption but also decreases the task completion score due to diminished algorithm detection capabilities.
To enhance the task completion score, we could reduce flight speed, employ a detection model with higher complexity, or increase the sampling rate; however, these approaches also raise energy consumption.
Therefore, selecting appropriate flight parameters to meet task requirements with energy efficiency poses a challenge.

(2) A large search space of flight parameters needs to be compressed.
Assuming there are nine altitude options (ranging from 20 to 100 meters in 10-meter increments), five speed options (ranging from 1 to 9$m/s$ in 2$m/s$ increments), four model options, and six sampling rate options for each model at each altitude and speed, we have 1080 sets of flight parameters to select from.
As the number of altitude, speed, model or sampling rate options increases, the choice of flight parameters expands exponentially.
Searching for the optimal trade-off between energy consumption and task completion score in such a vast search space is important and difficult.
Consequently, determining how to compress the search space to identify the most appropriate flight parameter among numerous options is challenging.

(3) Limited by the scarce computing resources of edge devices, the decision algorithm must be as lightweight as possible to ensure the operation of detection algorithms.

(4) Tolerance for energy consumption should be taken into account.
Within a tolerable range, if the energy consumption of flight parameter $f1$ is higher than that of $f2$, but the task completion score of $f1$ is greater than $f2$, considering $f1$ as the recommended parameter proves more beneficial for real-world tasks.

\textbf{Initially, we consider a scenario with only one detection model.}
The algorithm employs four experimentally verified conclusions as follows:

\textit{Conclusion 1:} The maximum task completion score does not increase with a rise in flight altitude.

\textit{Conclusion 2:} As flight altitude decreases, UAV energy consumption for detecting per square kilometer area increases.
However, within a specific altitude range, reducing the sampling rate of algorithms as the flight altitude decreases can compensate for the increased UAV energy consumption.

\textit{Conclusion 3:} System energy consumption declines when flight speed is increased.

\textit{Conclusion 4:} Provided the sampling rate is adjusted proportionally with speed, the task completion score remains unchanged as speed varies.

The proofs of these conclusions are presented in Section \ref{sec6_1}.

\begin{algorithm}[htb]
\caption{Single Model Decision Function}
\begin{algorithmic}[1] 
\REQUIRE ~~\\ 
    The detection model name, $name$; \\
    The UAV parameter, $\alpha$; \\
    The camera parameter, $\theta$; \\
    The flight altitude record from low to high, $H$; \\
    The maximum task completion score at each altitude, $perf_{max}$; \\
    The task completion score with different sampling rate at each altitude, $perf_{r}$; \\
    The minimum task completion score set by the user, $perf_{min}$; \\
    The running and standby energy consumption of the edge device, $P_{device}$; \\
    The maximum sampling rate of the model on the edge device, $r_{max}$; \\
    The altitude selection range, $n$; \\
    The sets of flight parameters to be recommended, $results$
\ENSURE ~~\\ 
    The sets of flight parameters to be recommended, $results$

\STATE $h_{max} \leftarrow -1$ \vspace{1mm}
\FOR{each $perf \in perf_{max}$} \vspace{1mm}
    \IF{$perf \ge perf_{min}$} \vspace{1mm}
        \STATE $h_{max} \leftarrow h_{max} + 1$ \vspace{1mm}
    \ELSE \vspace{1mm}
        \STATE break \vspace{1mm}
    \ENDIF \vspace{1mm}
\ENDFOR \vspace{1mm}

\FOR{each $i \in [h_{max} - n, h_{max}]$} \vspace{1mm}
    \FOR{each $(perf, r_s) \in perf_r[i]$} \vspace{1mm}
        \IF{$perf \ge perf_{min}$} \vspace{1mm}
            \STATE $v \leftarrow min(v_{max}, \frac{r_{max}}{r_s}v_s)$ \vspace{1mm}
            \STATE $r \leftarrow \frac{v}{v_s}r_s$ \vspace{1mm}
            \STATE $h \leftarrow H[i]$ \vspace{1mm}
            \STATE $p_{run}, P_{standby} \leftarrow P_{device}$ \vspace{1mm}
            \STATE $P_{system} \leftarrow EnergyCalculation()$ \vspace{1mm}
            \STATE $results.append([name, h, v, r, perf, P_{system}])$ \vspace{1mm}
        \ENDIF \vspace{1mm}
    \ENDFOR \vspace{1mm}
\ENDFOR \vspace{1mm}

\RETURN $results$ 
\end{algorithmic}
\label{alg_singer_model}
\end{algorithm}

\begin{algorithm}[htb]
\caption{Decision Function Based on Energy Efficiency}
\begin{algorithmic}[1]
\REQUIRE ~~\\
    The UAV parameter, $\alpha$; \\
    The camera parameter, $\theta$; \\
    The energy tolerance parameter, $\beta$; \\
    The flight altitude record from low to high, $H$; \\
    The altitude selection range, $n$; \\
    The minimum task completion score set by the user, $perf_{min}$; \\
    The different model performance, $data_{models}$
\ENSURE ~~\\
    The flight parameter to be recommended, $recommend$

\STATE $results \leftarrow list()$ \vspace{1mm}

\FOR{each $modelParameter \in data_{models}$} \vspace{1mm}
    \STATE $results \leftarrow Algorithm \ref{alg_singer_model}()$ \vspace{1mm}
\ENDFOR \vspace{1mm}

\STATE $energy_{min} \leftarrow minEnergy(results)$ \vspace{1mm}
\STATE $results \leftarrow remove(results, \beta \cdot energy_{min})$ \vspace{1mm}
\STATE $results \leftarrow seletMaxPerf(results)$ \vspace{1mm}
\STATE $recommend \leftarrow selectMinEnergy(results)$ \vspace{1mm}

\RETURN $recommend$ 
\end{algorithmic}
\label{alg_decision}
\end{algorithm}

Algorithm \ref{alg_singer_model} represents the single model decision function.
Lines 1 to 8 are used to obtain the maximum altitude that satisfies the task completion score.
According to \textit{Conclusion 1}, altitudes higher than the maximum altitude will not fulfill the task completion requirements.
The term $h_{max}$ denotes the subscript in the flight altitude record $H$.
By filtering out altitudes that do not meet the task completion requirement, we narrow the search space.
According to \textit{Conclusion 2}, we tend to select a higher altitude to reduce UAV energy consumption.
Meanwhile, within a specific range, we can decrease the altitude as the sampling rate decreases to find suitable flight parameters without increasing system energy consumption.
Therefore, line 9 narrows the search space from $h_{max} - n$ to $h_{max}$.
Different devices have varying altitude selection range values $n$.
The example of the calculation of $n$ will be presented in Section \ref{sec5_energy}.
For each altitude, different sampling rates $r_s$ yield different task completion scores.
Lines 10 and 11 are employed to identify all parameters that satisfy the task completion requirement.
In accordance with \textit{Conclusion 3}, we prefer the fastest possible speed to minimize energy consumption.
As the speed increases, we need to raise the sampling rate to ensure that the task completion score remains unchanged, as per \textit{Conclusion 4}.
Two factors limit speed increase: the maximum speed of the UAV platform and the maximum sampling rate of the edge device.
Line 12 determines the maximum speed that the system can achieve.
Under the benchmark speed $v_s$, the UAV achieves a task completion score of $perf$ with a sampling rate of $r_s$.
Consequently, the maximum flight speed is $(r_{max} / r_s) \times v_s$ under the limitation of the edge device, maintaining the same task completion score $perf$.
The maximum achievable speed $v$ of the system is the minimum value between the maximum speed $v_{max}$ of the UAV platform and $(r_{max} / r_s) \times v_s$.
Line 13 then calculates the sampling rate based on the speed $v$.
Line 14 records the current altitude, and line 15 obtains the running and standby energy consumption of the detection algorithm.
Using the aforementioned data, line 16 computes the system energy consumption employing the energy consumption calculation function based on Formula \ref{formula_system_power}.
Line 17 stores the flight parameters along with their energy consumption.
In summary, Algorithm \ref{alg_singer_model} saves all flight parameters for a detection model that meets the task completion requirement.

\textbf{Subsequently, we introduce the decision-making process for multiple models, as illustrated in Algorithm \ref{alg_decision}.}
Lines 1 to 4 employ Algorithm \ref{alg_singer_model} to obtain all potential sets of flight parameters.
Line 5 records the minimum system energy consumption associated with these flight parameters.
Subsequently, line 6 filters out the flight parameters with energy consumption exceeding $\beta \cdot energy_{min}$, where $\beta$ represents the energy tolerance parameter.
This parameter serves to prevent the exclusion of flight parameters that exhibit significantly improved detection performance and marginally higher energy consumption compared to those with minimum energy consumption.
In line 7, the flight parameters with the highest task completion scores are selected from the filtered parameters.
If multiple flight parameters share the best task completion score, line 8 chooses the parameter with the lowest energy consumption as the recommended option.

Algorithm \ref{alg_decision} enables the identification of the most appropriate flight parameter.
\textbf{This approach offers several advantages}, such as (1) leveraging the appropriate decision order to circumvent the mutual interference between flight altitude, speed, sampling rate, and detection model, (2) compressing the search space by adhering to flight rules derived from experimental data, thereby substantially reducing computational requirements, and (3) incorporating the energy tolerance parameter to strike a balance between energy consumption and task completion score, ensuring that superior flight parameters are not overlooked.

\section{Experiment}
In this section, we first present the details of our experimental environment and evaluate the performance of our object detection models.
Following this, we utilize actual data to measure the task completion score, evaluate the performance of our energy efficiency-oriented decision algorithm, and assess the energy consumption of the system.
After analyzing the measurement results, we extract the relevant system parameters.
Finally, we evaluate the performance of the system in real-world scenarios.

\begin{table*}[ht]
\centering
\caption{The precision and recall of anchor box optimization in the VisDrone dataset.}
\begin{tabular}{cccccccc}
\toprule
\textbf{} & \textbf{AP ($\%$)} & \textbf{AP50 ($\%$)} & \textbf{AP75 ($\%$)} & \textbf{AR1($\%$)} & \textbf{AR10 ($\%$)} & \textbf{AR100 ($\%$)} & \textbf{AR500 ($\%$)} \\
\midrule
\textbf{Basic method} & 4.46 & 11.85 & 2.46 & 0.11 & 1.4 & 8.68 & 9.85 \\
\textbf{Optimization method} & 8 & 18.39 & 6.02 & 0.34 & 2.18 & 11.41 & 14.48
\\
\toprule
\end{tabular}
\label{table_anchor_optimization}
\end{table*}

\subsection{Experimental environment}
The computational hardware utilized for model training consists of an Intel Xeon E5-2620v4 CPU and an NVIDIA Quadro P5000/16GB GPU.
The edge device employed is the NVIDIA Jetson Xavier NX, configured with NumPy 1.16.0 and TensorRT 7.0 software versions.
The UAV platform utilized in this study is the DJI Mavic Air 2.

\subsection{Model performance}
Section \ref{sec3_1} presents the anchor box optimization method.
For the VisDrone dataset, anchor boxes were optimized from [(1, 2), (2, 3), (2, 5), (5, 4), (4, 7), (9, 6), (6, 12), (15, 11), (25, 25)] to [(2, 4), (4, 6), (4, 10), (10, 8), (8, 14), (18, 12), (12, 24), (30, 22), (50, 50)] with a $416 \times 416$ input resolution.
The optimization performance is demonstrated in Table \ref{table_anchor_optimization}.
It is evident that the average precision (AP) and average recall (AR) are substantially improved for the VisDrone dataset following optimization, thereby confirming the viability of the optimization method.
The detection performance of various models in practical tasks will be discussed in Section \ref{sec5_task_completion}.

\begin{algorithm}[htb]
\caption{Sampling Rate Search Function}
\begin{algorithmic}[1] 
\REQUIRE ~~\\ 
    The initial sampling rate, $r1$; \\
    The sampling rate tolerance parameter, $\epsilon$; \\
\ENSURE ~~\\ 
    The most reasonable sampling rate for the maximum E$^3$-UAV recall.

\STATE $r2 \leftarrow r1 * 2$ \vspace{1mm}

\WHILE{$Recall(r2) > Recall(r1)$} \vspace{1mm}
    \STATE $r1 \leftarrow r2$ \vspace{1mm}
    \STATE $r2 \leftarrow r1 * 2$ \vspace{1mm}
\ENDWHILE \vspace{1mm}

\STATE $r_{top} \leftarrow r1$ \vspace{1mm}
\STATE $r_{bottom} \leftarrow \frac{r1}{2}$ \vspace{1mm}

\WHILE{$r_{top} - r_{bottom} < \epsilon$} \vspace{1mm}
    \IF{$Recall(\frac{r_{top} + r_{bottom}}{2}) < Recall(r_{top})$} \vspace{1mm}
        \STATE $r_{bottom} \leftarrow \frac{r_{top} + r_{bottom}}{2}$ \vspace{1mm}
    \ELSE \vspace{1mm}
        \STATE $r_{top} \leftarrow \frac{r_{top} + r_{bottom}}{2}$ \vspace{1mm}
    \ENDIF \vspace{1mm}
\ENDWHILE \vspace{1mm}

\RETURN $r_{top}$ 
\end{algorithmic}
\label{alg_sampling_rate}
\end{algorithm}

\subsection{Task completion score and analysis}\label{sec5_task_completion}

In this study, we focus on the detection of two prevalent object types: person and car.
The E$^3$-UAV recall is employed to represent the task completion score, as numerous tasks prioritize recall.
For instance, in search and rescue operations, it is preferable to identify a larger number of objects.

\textbf{A primary challenge lies in determining the appropriate sampling rate.}
For car detection, an excessively high sampling rate increases energy consumption due to increased computation, while an overly low rate hinders the E$^3$-UAV from achieving optimal detection performance.
To obtain the appropriate sampling rate, the following steps are presented.

(1) Selection of the initial sampling rate $r1$: The first frame in the video is analyzed, yielding the distance of the farthest object detected by the model. Subsequently, the time $\Delta t$ required to fly over the object is calculated. The initial sampling rate is determined as $1/\Delta t$, which represents object detection at intervals of $\Delta t$. Ideally, the UAV should detect all objects at a $1/\Delta t$ sampling rate, however, actual environmental conditions may prevent the detection of numerous objects. Consequently, it is crucial to reasonably increase the sampling rate to enhance detection performance.

\begin{figure}[ht]
\centering
\includegraphics[width=2.5in]{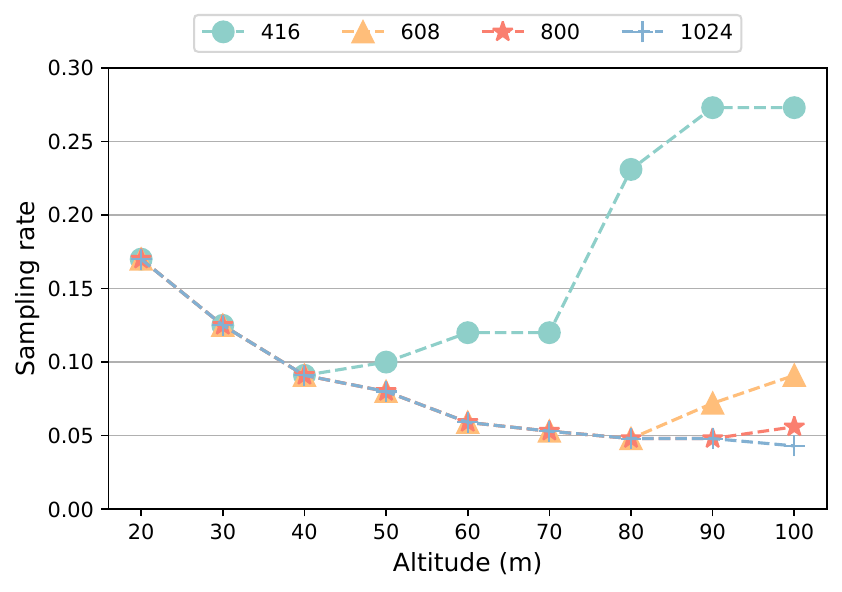}
\caption{The sampling rate across different altitudes and models when the E$^3$-UAV recall of car detection reaches $100\%$. The 416, 608, 800, and 1024 represent the models whose input resolution is $416 \times 416$, $608 \times 608$, $800 \times 800$, and $1024 \times 1024$, respectively.}
\label{fig_car_sampling_rate}
\end{figure}

(2) Algorithm \ref{alg_sampling_rate} employs a binary search method to identify the most reasonable sampling rate that maximizes the E$^3$-UAV recall.
As illustrated in lines 1 to 5, if the E$^3$-UAV recall does not reach its maximum, the sampling rate will be doubled continuously until the maximum E$^3$-UAV recall is attained.
At this point, $r1$ is set to $r_{top}$ and $r1/2$ is set to $r_{bottom}$.
The most appropriate sampling rate lies between $r_{bottom}$ and $r_{top}$.
Lines 8 to 14 repeatedly reduce the search space by half until the search range is less than $\epsilon$.
Subsequently, $r_{top}$ corresponds to the desired sampling rate.

This method offers three advantages: (1) the rapid identification of the sampling rate that maximizes the E$^3$-UAV recall, (2) the utilization of the sampling rate tolerance parameter $\epsilon$ to reduce the number of searches while avoiding excessively high sampling rates, and (3) the generation of multiple sampling rate and E$^3$-UAV recall sets during the search, enriching the selection of the system in sampling rate, thereby assisting the system in selecting the most reasonable sampling rate in accordance with task requirements.

\textbf{For car detection}, we measure the maximum E$^3$-UAV recall across various models and altitudes, considering a speed of $5m/s$.
Our findings indicate that the maximum E$^3$-UAV recall reaches $100\%$ for all models and altitudes, as cars are relatively simple objects to detect.
Figure \ref{fig_car_sampling_rate} depicts the sampling rates for different altitudes and models when the E$^3$-UAV recall for car detection attains $100\%$. Several observations can be drawn from these results:

(1) As the input resolution increases, the sampling rate decreases, suggesting that a larger input resolution yields superior detection performance.

(2) With increasing altitude, the sampling rate initially declines before rising again.
Models with smaller input resolutions exhibit a lower ascending node altitude.
This trend can be attributed to the expanding detection area and the increased number of detected objects as the altitude rises, which allows for a reduction in sampling rate without object omission.
However, upon reaching a specific altitude, object size diminishes, and the detection capabilities of models gradually decline with increasing altitude.
Consequently, models struggle to accurately detect distant objects, necessitating an elevation in the sampling rate to enhance the E$^3$-UAV recall.

\begin{figure}[ht]
\centering
\includegraphics[width=2.5in]{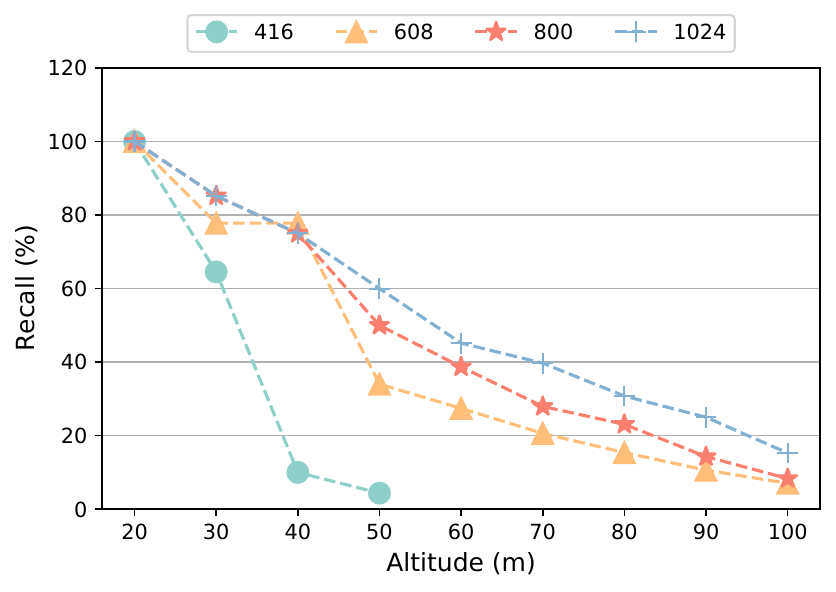}
\caption{The maximum E$^3$-UAV recall ($\%$) across different altitudes and models for person detection. The 416, 608, 800, and 1024 represent the models whose input resolution is $416 \times 416$, $608 \times 608$, $800 \times 800$, and $1024 \times 1024$, respectively.}
\label{fig_person_recall}
\end{figure}

\begin{figure}[ht]
\centering
\includegraphics[width=2.5in]{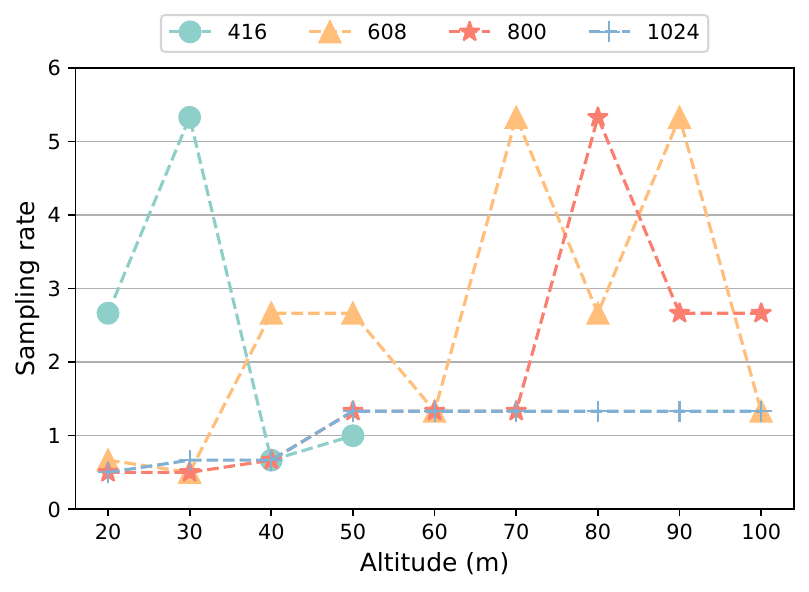}
\caption{The sampling rate across different altitudes and models when the E$^3$-UAV recall of person detection reaches a maximum. The 416, 608, 800, and 1024 represent the models whose input resolution is $416 \times 416$, $608 \times 608$, $800 \times 800$, and $1024 \times 1024$, respectively.}
\label{fig_person_sampling_rate}
\end{figure}

\textbf{For person detection}, the sampling rate search method is analogous to that of car detection.
The primary distinction lies in the initial sampling rate, which is set at 0.333 for all models and altitudes.
This is due to the small size of person objects, resulting in a limited effective detection distance.
Thus images with different object densities can yield varying initial sampling rates for a specific model at a given altitude, rendering it challenging to accurately measure the initial sampling rate across diverse scenes.

Figure \ref{fig_person_recall} presents the maximum E$^3$-UAV recall for person detection across various altitudes and models.
The maximum E$^3$-UAV recall increases with the input resolution, signifying that larger input resolutions deliver superior detection performance for small objects.
The maximum E$^3$-UAV recall diminishes significantly with increasing altitude, highlighting the importance of altitude as a critical parameter affecting the overall performance of models for person detection.
Notably, the $416 \times 416$ model is incapable of detecting person objects when the altitude exceeds 60 meters.
These observations suggest that person detection serves as a challenging benchmark to assess the capabilities of the system.

\begin{table}[ht]
\centering
\caption{The average power and latency of the decision algorithm on the Jetson Xavier NX across various task requirements.}
\begin{tabular}{ccc}
\toprule
\textbf{Task completion score ($\%$)} & \textbf{Avg. Power ($W$)} & \textbf{Avg. Latency ($ms$)} \\
\midrule
20 & 7.1 & 0.062 \\
30 & 7.2 & 0.075 \\
40 & 7 & 0.076 \\
50 & 7.1 & 0.065 \\
66.67 & 7.5 & 0.054 \\
\bottomrule
\end{tabular}
\label{table_decision_average_power}
\end{table}

\begin{table}[ht]
\centering
\caption{The energy consumption per second ($Wh/s$) of Jetson Xavier NX across different models and frameworks.}
\begin{tabular}{ccccc}
\toprule
\textbf{Model} & \textbf{TensorFlow} & \textbf{PyTorch} & \textbf{TensorRT} & \textbf{TensorRT-16bit} \\
\midrule
416 & 0.00523 & 0.00558 & 0.00442 & 0.00241 \\
608 & 0.00524 & 0.00494 & 0.00419 & 0.00273 \\
800 & 0.00529 & 0.00487 & 0.00437 & 0.00298 \\
1024 & 0.00537 & 0.00528 & 0.00369 & 0.00257
\\
\toprule
\end{tabular}
\label{table_edge_energy}
\end{table}

Figure \ref{fig_person_sampling_rate} displays the sampling rates for different altitudes and models when the E$^3$-UAV recall for person detection reaches its maximum.
The sampling rate is not recorded for the $416 \times 416$ input resolution model when the altitude exceeds 60 meters, as it fails to detect person objects.
The fluctuations in the sampling rate with increasing altitude are significant for the $608 \times 608$ and $800 \times 800$ input resolution models, primarily due to their limited detection capabilities.
Models with restricted detection abilities exhibit unstable performance in diverse scenes at varying altitudes.
In contrast, the sampling rate remains stable as the altitude increases for the $1024 \times 1024$ input resolution model, as this model outperforms the others in terms of detection performance.

\subsection{Performance of energy efficiency-oriented decision algorithm}

We evaluate the average power consumption and latency of the decision algorithm on the Jetson Xavier NX under varying task completion score thresholds, as illustrated in Table \ref{table_decision_average_power}.
The average power consumption of the algorithm exceeds the standby power of $6W$ by slightly over $1W$.
Meanwhile, the average latency remains below $0.1ms$.
These findings demonstrate that the decision algorithm minimally impacts the operation of other functions and exhibits modest energy consumption.

\subsection{System energy consumption calculation and analysis}\label{sec5_energy}

In this subsection, we calculate the system parameters to assess and analyze the system energy consumption.

\begin{figure}[ht]
\centering
\includegraphics[width=2.0in]{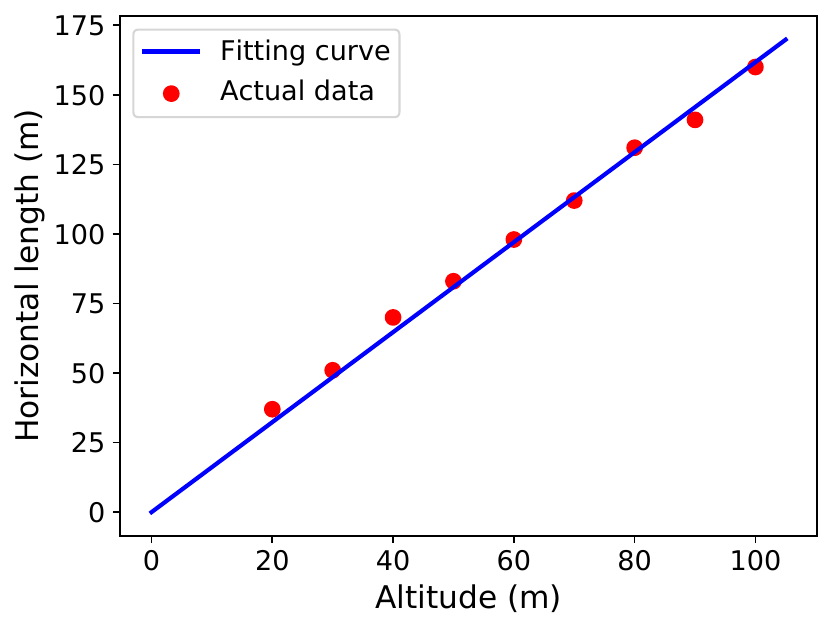}
\caption{
The horizontal length $l$ of the detection area varies with altitude. The red points signify the empirical data obtained from real-world measurements, while the blue curve represents the fitting curve derived using the Least Squares method. The camera parameter $\theta$ is determined to be 1.617.}
\label{fig_horizontal_length}
\end{figure}

\textbf{Initially, we calculate the camera parameter $\theta$.}
we measure the horizontal length $l$ of the detection area at different altitudes $h$, employing ground markers for measurement.
As errors are inevitable in practical measurements, we apply the Least Squares method to fit the collected data according to the relationship $l = \theta h$, as depicted in Fig. \ref{fig_horizontal_length}.
The fitted camera parameter, $\theta$, is found to be 1.617.

\begin{figure}[ht]
\centering
\includegraphics[width=2.0in]{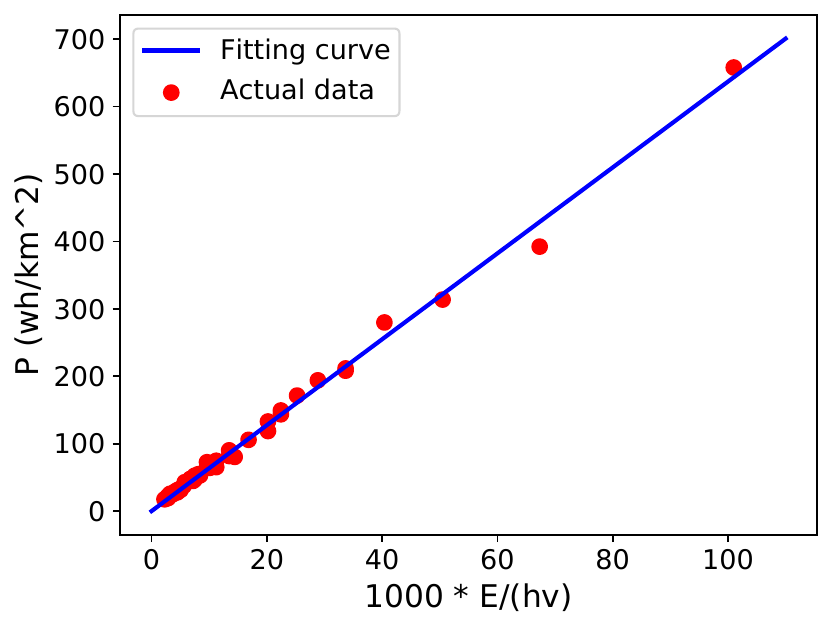}
\caption{
The fitting of the UAV parameter is conducted in accordance with Formula \ref{formula_uav_power}. The red points indicate the actual data, while the blue curve represents the fitted curve. Due to its small magnitude, the $E/(hv)$ value is magnified 1000 times for clearer visualization. Employing the Least Squares method, the value of $1 / \alpha$ is determined to be 6368.874, and the value of $\alpha$ is $1.57 \times 10^{-4}$}
\label{fig_uav_parameter}
\end{figure}

\textbf{Subsequently, we calculate the UAV parameter $\alpha$.}
we measure the flight time of the UAV at various altitudes and speeds, given an unit energy consumption $E$ of $2.02Wh$.
Altitudes are measured in increments of 10 meters, ranging from 20 to 100 meters, while speeds span from 1 to $9m/s$, increasing in $2m/s$ intervals.
Then we compute the detection area $S$ at varying altitudes and speeds based on the flight time $t$, horizontal length $l$, and speed $v$.
According to $P_{UAV} = E / S$, we calculate the UAV energy consumption $P_{UAV}$ at different altitudes and speeds.
Figure \ref{fig_uav_parameter} illustrates the application of the Least Squares method to fit the aforementioned data, obtaining the reciprocal of the UAV parameter $1 / \alpha$ using Formula \ref{formula_uav_power}.
The value of $1 / \alpha$ is determined to be 6368.874, and the value of $\alpha$ is $1.57 \times 10^{-4}$.

\textbf{For edge energy consumption}, Table \ref{table_edge_energy} presents the energy consumption per second of the Jetson Xavier NX when utilizing various models and frameworks.
The standby energy consumption of the device is $0.00168Wh/s$.
We examine the trade-off between UAV and edge energy consumption within a specific context.
Formula \ref{formula_analysis_1} outlines the UAV energy consumption as a function of $E$ and $\alpha$.
To calculate the running energy consumption, we employ the highest edge running energy consumption of $0.00537Wh/s$, associated with the TensorFlow framework and an input resolution of $416 \times 416$.
Formula \ref{formula_analysis_2} defines the edge device running energy consumption based on $\theta$.
Formula \ref{formula_analysis_3} describes the standby energy consumption when the edge device is not executing the algorithm.
Formula \ref{formula_analysis_4} delineates the system energy consumption during the operation of the algorithm with the highest edge running energy consumption.
Formula \ref{formula_analysis_5} illustrates the system energy consumption in standby mode.

\textbf{From the aforementioned formulas, we can determine the altitude selection range $n$.}
System energy consumption is expressed as $P = k / hv$, which represents an inverse proportionality.
According to Formula \ref{formula_analysis_4}, the running energy consumption is $16.787Wh/km^2$ at the maximum flight altitude of $100m$ and maximum speed of $10m/s$.
We then consider a specific scenario in which the altitude is decreased to enhance energy consumption, and standby mode is utilized to reduce energy consumption.
Based on Formula \ref{formula_analysis_5}, the flight altitude is $82m$ when the standby energy consumption equals the aforementioned running energy consumption.
Standby energy consumption will be lower than running energy consumption when the altitude exceeds $82m$.
Thus, the flight parameter trade-off holds significant value within the altitude interval between $82m$ and $100m$.
Additionally, decreasing the altitude will narrow the altitude interval in accordance with the features of inverse proportionality.
Based on these calculations, it is appropriate to select an altitude range $n$ of 2 for the experiment, covering a range of 30 meters.

\begin{equation}
P_{UAV} = \frac{E}{\alpha hv} = \frac{12866.242}{hv}
\label{formula_analysis_1}
\end{equation}

\begin{equation}
P_{edge1} = \frac{10^6}{\theta hv}P_{run} = \frac{3320.965}{hv}
\label{formula_analysis_2}
\end{equation}

\begin{equation}
P_{edge2} = \frac{10^6}{\theta hv}P_{standby} = \frac{1038.961}{hv}
\label{formula_analysis_3}
\end{equation}

\begin{equation}
P_{UAV} + P_{edge1} = \frac{16187.207}{hv}
\label{formula_analysis_4}
\end{equation}

\begin{equation}
P_{UAV} + P_{edge2} = \frac{13905.203}{hv}
\label{formula_analysis_5}
\end{equation}

The aforementioned calculations demonstrate the impact of altitude on the sampling rate.
\textbf{The following analysis focuses on the influence of speed.}
The sampling rate exhibits a proportional increase with rising speed.
To simplify the calculation, we assume that the maximum sampling rate is ten and that the speed is equal to the sampling rate.
Formula \ref{formula_analysis_speed} illustrates the edge device energy consumption at an altitude of 100 meters, expressed in the form $P = a + b/v$.
While the values of $a$ and $b$ may change when adjusting the maximum sampling rate and the proportionality between the sampling rate and speed, the structure of the formula remains constant.
This suggests that the edge energy consumption decreases as speed increases.
Consequently, selecting a higher speed is preferred to reduce energy consumption.

\begin{equation}
\begin{aligned}
P_{edge3} &= \frac{10^6r}{\theta hvr_{max}}P_{run} + (1 - \frac{r}{r_{max}})\frac{10^6}{\theta hv}P_{standby}\\
&= 2.282 + \frac{10.39}{v}
\label{formula_analysis_speed}
\end{aligned}
\end{equation}

\subsection{System performance}\label{sec5_perf}

Based on the preceding experiments, we derive the system parameters, which are presented in Table \ref{table_system_parameter}.
The energy consumption of various models introduced in Section \ref{sec3_1} within the NVIDIA Jetson Xavier NX is shown in Table \ref{table_NX_performance}.

\begin{table}[ht]
\centering
\caption{System parameter.}
\begin{tabular}{cc}
\toprule
\textbf{Parameter} & \textbf{Value} \\
\midrule
The unit energy consumption $E$ & 2.02 \\
The UAV parameter $\alpha$ & $1.57 \times 10^{-4}$ \\
The camera parameter $\theta$ & 1.617 \\
The altitude record $H$ & [20, 30, ..., 100] \\
The energy tolerance parameter $\beta$ & 1.05 \\
The altitude selection range $n$ & 2 \\
The edge energy consumption $P_{device}$ & See Table \ref{table_NX_performance}\\
The maximum sampling rate $r_{max}$ & See Table \ref{table_NX_performance}
\\
\toprule
\end{tabular}
\label{table_system_parameter}
\end{table}

\begin{table}[ht]
\centering
\caption{The maximum sampling rate and energy consumption of the NVIDIA Jetson Xavier NX across the different models.}
\begin{tabular}{ccc}
\toprule
\textbf{Model} & \textbf{$r_{max}$} & \textbf{Energy consumption $(Wh/s)$} \\
\midrule
416 & 18.232 & 0.00241 \\
608 & 6.379 & 0.00273 \\
800 & 3.921 & 0.00298 \\
1024 & 1.749 & 0.00257 \\
standby & / & 0.00168
\\
\toprule
\end{tabular}
\label{table_NX_performance}
\end{table}

\begin{table*}[ht]
\centering
\caption{The car detection performance.}
\scalebox{0.85}{
\begin{tabular}{cccccccc}
\toprule
\textbf{} & \textbf{Object} & \textbf{Altitude $(m)$} & \textbf{Speed $(m/s)$} & \textbf{Sampling rate} & \textbf{Model} & \textbf{E$^3$-UAV recall $(\%)$} & \textbf{Energy consumption $(Wh)$} \\
\midrule
\textbf{E$^3$-UAV} & \textbf{car} & \textbf{100} & \textbf{10} & \textbf{0.086} & \textbf{1024} & \textbf{93.33} & \textbf{4.269} \\
Adjacent & car & 100 & 10 & 1.749 & 1024 & 100 & 4.382 \\
Adjacent & car & 100 & 10 & 0.086 & 800 & 71.11 & 4.267 \\
Adjacent & car & 90 & 10 & 0.086 & 1024 & 95.56 & 5.113\\
Adjacent & car & 100 & 8 & 0.086 & 1024 & 97.78 & 5.148
\\
\toprule
\end{tabular}}
\label{table_car_performance}
\end{table*}

\textbf{Initially, we measured the car detection performance and utilized adjacent parameters to validate the energy efficiency of E$^3$-UAV.}
We established an E$^3$-UAV recall requirement of $80\%$.
The recommended flight parameters are a $100m$ flight altitude, $10m/s$ flight speed, $0.086$ sampling rate, and a $1024 \times 1024$ input resolution model.
Table \ref{table_car_performance} presents the detection performance.
Line 1 displays the performance for the recommended parameters, achieving a $93.33\%$ E$^3$-UAV recall and a system energy consumption of $4.269 Wh$.
We then compare the performance of adjacent flight parameters, maintaining the same detection area.
Line 2 depicts the performance with the maximum sampling rate.
It reveals an improvement in E$^3$-UAV recall and an increase in energy consumption compared to the recommended parameters.
However, the energy consumption of 4.382 is not greater than $4.269 \cdot \beta$, which contradicts the decision-making strategy.
This discrepancy primarily arises from the reality that the collected data's scenes fail to encapsulate the entirety of actual scenarios; nonetheless, the system determines the sampling rate predicated upon user-collected data.
Consequently, we risk overlooking more favorable alternatives when data collection is restricted to a narrow scope.
We thoroughly examine the issue and elucidate strategies for enhancing performance in Section \ref{sec6_2}, \textit{Insight 2}.
Line 3 demonstrates the performance with an $800 \times 800$ input resolution model.
The E$^3$-UAV recall fails to meet the task requirement, and the energy consumption is nearly identical to that of the recommended parameters.
Line 4 exhibits the performance at $90m$ altitude, wherein the E$^3$-UAV recall marginally increases, but the energy consumption notably rises compared to the recommended parameters.
Line 5 illustrates the performance with an $8m/s$ speed, also displaying a slight increase in E$^3$-UAV recall and a significant increase in energy consumption compared to the recommended parameters.
In conclusion, the system can satisfy the task requirements with exceptional energy efficiency for car detection.

\begin{table*}[ht]
\centering
\caption{The person detection performance.}
\scalebox{0.85}{
\begin{tabular}{cccccccc}
\toprule
\textbf{} & \textbf{object} & \textbf{altitude $(m)$} & \textbf{speed $(m/s)$} & \textbf{sampling rate} & \textbf{model} & \textbf{E$^3$-UAV recall $(\%)$} & \textbf{energy consumption $(Wh)$} \\
\midrule
\textbf{E$^3$-UAV} & \textbf{person} & \textbf{50} & \textbf{10} & \textbf{1.332} & \textbf{1024} & \textbf{61.45} & \textbf{2.263} \\
Adjacent & person & 50 & 10 & 1.749 & 1024 & 61.74 & 2.285 \\
Adjacent & person & 50 & 10 & 1.332 & 800 & 48.99 & 2.239 \\
Adjacent & person & 40 & 10 & 1.332 & 1024 & 74.5 & 3.085 \\
Adjacent & person & 60 & 10 & 1.332 & 1024 & 43.85 & 1.816 \\
Adjacent & person & 50 & 8 & 1.332 & 1024 & 62.48 & 2.672
\\
\toprule
\end{tabular}}
\label{table_person_performance}
\end{table*}

\begin{table*}[ht]
\centering
\caption{The performance of random flights and the recommended flight parameter for person detection.}
\scalebox{0.85}{
\begin{tabular}{cccccccc}
\toprule
\textbf{} & \textbf{object} & \textbf{altitude $(m)$} & \textbf{speed $(m/s)$} & \textbf{sampling rate} & \textbf{model} & \textbf{E$^3$-UAV recall $(\%)$} & \textbf{energy consumption $(Wh)$} \\
\midrule
Baseline & person & random & random & 18.232 & 416 & 17.56 & 3.083 \\
Baseline & person & random & random & 6.379 & 608 & 55.76 & 3.117 \\
Baseline & person & random & random & 3.921 & 800 & 62.34 & 3.144 \\
Baseline & person & random & random & 1.749 & 1024 & 66.67 & 3.100 \\
\textbf{E$^3$-UAV} & \textbf{person} & \textbf{40} & \textbf{10} & \textbf{5.334} & \textbf{608} & \textbf{74.04} & \textbf{3.107}
\\
\toprule
\end{tabular}}
\label{table_random_performance}
\end{table*}

\textbf{Subsequently, we measured the person detection performance.}
We established a $50\%$ E$^3$-UAV recall requirement for person detection.
The recommended flight parameters include a $50m$ flight altitude, $10m/s$ flight speed, $0.086$ sampling rate, and a $1024 \times 1024$ input resolution model.
The E$^3$-UAV recall and energy consumption for the recommended flight parameters are $61.45\%$ and $2.263Wh$, respectively, as displayed in Table \ref{table_person_performance}, line 1.
We then compared the performance of adjacent flight parameters individually.
Line 2 exhibits a better E$^3$-UAV recall within the energy tolerance range, however, the system does not recommend it.
This outcome is due to the data not covering all possible selections, similar to the case with car detection.
The energy consumption of lines 3 and 5 is less than $2.263Wh$, but they fail to meet the task requirements.
Lines 4 and 6 present energy consumption values that surpass the energy tolerance range.
These findings indicate that the system is adept at selecting suitable parameters for person detection tasks while maintaining a great energy efficiency.

\textbf{Furthermore, we employed random flight as a baseline to verify the energy efficiency of E$^3$-UAV.}
The performance of random flight and the recommended flight parameters are presented in Table \ref{table_random_performance}.
In lines 1 to 4, we utilize four models with the highest sampling rate to detect persons, integrating random variations in altitude and speed.
Subsequently, we select the maximum E$^3$-UAV recall ($66.67\%$) from these models as the task requirement.
Line 5 displays the recommended flight parameter and its performance.
Although the energy consumption is nearly identical to the others, the E$^3$-UAV recall demonstrates a significant improvement, reaching $74.04\%$.
This result signifies that the system can enhance detection performance compared to manual random flight.

At present, only a limited number of systems permit the trade-off of multiple parameters during object detection tasks.
Li \textit{et al.}~\cite{sys_3_li2019intelligent} and Kaliappan~\textit{et al.}~\cite{sys_5_madasamy2021osddy} constructed two actual object detection systems that do not optimize energy consumption through modeling and solely deploy detection algorithms on edges.
Lines 1 to 4 of Table \ref{table_random_performance} can represent the running mode of the aforementioned systems.
The E$^3$-UAV system achieves up to an $11.7\%$ reduction in edge energy consumption (from $0.316Wh$ to $0.279Wh$) and increases the recall rate from $62.34\%$ (line 3) to $74.04\%$ when compared to the systems mentioned above, as they do not incorporate any flight energy optimization techniques.

\section{Discussion}
In this section, we prove the conclusions used in previous sections, and introduce four insights to aid researchers in further studying UAV-based object detection.

\subsection{Proof of conclusions}\label{sec6_1}



\begin{figure}[ht]
\centering
\includegraphics[width=3.4in]{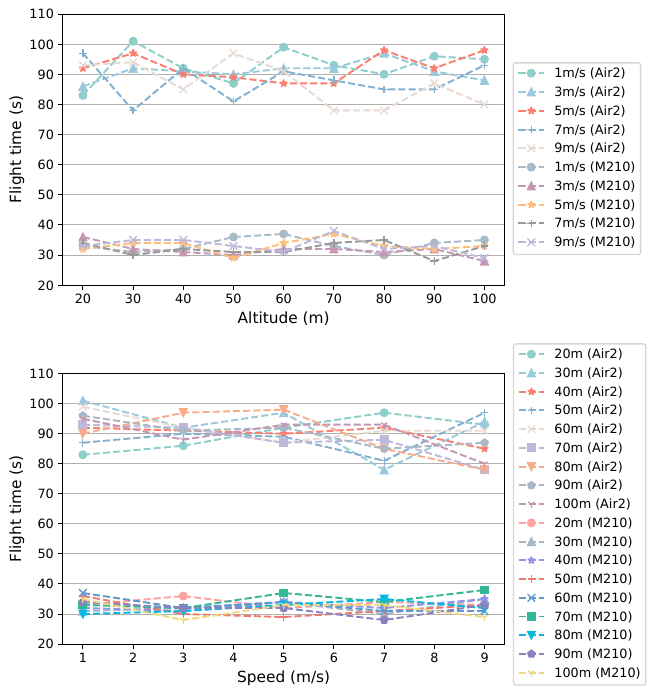}
\caption{The flight time with increase in altitude (top) and speed (bottom) at $2.02Wh$ energy consumption using DJI Mavic Air 2 and $6.984Wh$ energy consumption using DJI Matrice M210.}
\label{figure_flight_time}
\end{figure}

First, we prove that \textit{the influence of speed and altitude on average flight time per watt-hour can be disregarded in actual flights}, which is used in Section {\ref{sec3_3}}.
\textbf{Proof:} Figure \ref{figure_flight_time} illustrates the flight time across varying altitudes and speeds at an energy consumption of $2.02Wh$ using the DJI Mavic Air 2 and $6.984Wh$ using the DJI Matrice M210.
We observe that the flight time does not change consistently but remains stable within an interval as altitudes increase at different speeds. Furthermore, with the increase in speed, the flight time exhibits the same pattern at varying altitudes.
The standard deviation coefficient of flight time is $6.23\%$ for DJI Mavic Air 2.
If the flight time were to display a specific trend with the increase of altitude or speed, the standard deviation coefficient would not be so small.
For instance, the standard deviation coefficient of an arithmetic progression $[1, 2, 3, ... , 10]$ is $55.05\%$, and for a geometric sequence $[1, 2, 4, ..., 512]$, it is $161.34\%$.
Additionally, the actual data in Fig. \ref{fig_uav_parameter} is linear, which validates the correctness of Formula \ref{formula_uav_power}, which is based on this conclusion.
Thus, the actual data also supports the conclusion.
The standard deviation coefficient of flight time is $7.06\%$ for DJI Matrice M210, indicating that the conclusion is generalizable.
According to these empirical data, we can assume that the influence of speed and altitude on average flight time per watt-hour can be disregarded in actual flights, and the fluctuation of flight time is attributed to unpredictable variables such as wind speed, wind direction, and operation methods during each flight.
Theoretically, altitude and speed would affect air resistance, but their influence is not as significant compared to other unpredictable variables. We discuss this in \textit{Insight 1} below.

Next, we provide proofs for Conclusions 1 to 4 used in Section \ref{sec4_2}.

\textit{Conclusion 1:} The maximum task completion score does not increase with a rise in flight altitude.
\textbf{Proof:} Figure \ref{fig_person_recall} demonstrates that the maximum E$^3$-UAV recall for person detection declines as the flight altitude increases for all models.
The maximum E$^3$-UAV recall for car detection is $100\%$ for all models. Additionally, the maximum E$^3$-UAV precision decreases with the rising flight altitude in our experiments.
The aforementioned evidence supports \textit{Conclusion 1}.
As the flight altitude increases, objects appear smaller, and their quantity increases.
Theoretically, model detection performance will deteriorate with increasing flight altitude since objects become more challenging to detect.
\textit{Conclusion 1} is also intuitive.

\textit{Conclusion 2:} As flight altitude decreases, UAV energy consumption for detecting per square kilometer area increases.
However, within a specific altitude range, reducing the sampling rate of algorithms as the flight altitude decreases can compensate for the increased UAV energy consumption.
\textbf{Proof:} Section \ref{sec5_energy} calculate the UAV and edge energy consumption to get the altitude selection range $n$.
The calculation process proves \textit{Conclusion 2}.

\textit{Conclusion 3:} System energy consumption declines when flight speed is increased.
\textbf{Proof:} Formula \ref{formula_uav_power} indicates that the UAV energy consumption will decrease with the increased flight speed.
The analysis of Formula \ref{formula_analysis_5} indicates that the edge device energy consumption will also decrease with the increased flight speed.
In conclusion, the system energy consumption will decrease when increasing flight speed.

\textit{Conclusion 4:} Provided the sampling rate is adjusted proportionally with speed, the task completion score remains unchanged as speed varies.
\textbf{Proof:} Suppose a UAV detects an area at a speed of $v1$ and a sampling rate of $r1$, capturing $n$ images with a flight time of $t1$.
If the speed is altered to $s \times v1$, the flight time will change to $t1 / s$ while maintaining the same flight path.
By adjusting the sampling rate to $s \times r1$, we can capture the same $n$ images.
Since the same images are detected, the detection results will remain unchanged.
Consequently, the task completion score will not be affected.

\subsection{Insights}\label{sec6_2}

\textit{Insight 1:} Flight speed and altitude have minimal impact on the flight time per watt-hour for quadrotor civilian drones in actual flights.
This is because the flight altitude of quadrotor civilian drones is generally below 120 meters, and flight environments, such as air resistance, do not vary significantly within this altitude range.
Although air resistance is proportional to the square of the speed, the impact on motor power is negligible for quadrotor civilian drones, as their flight speeds are generally below $10m/s$.
Additionally, unpredictable variables, such as wind speed, wind direction, and operation methods in each flight, overshadow the slight influence of altitude and speed.
Therefore, we can disregard the impact of altitude and speed on flight time per watt-hour in actual flights.

\textit{Insight 2:} We found that although the decision algorithm can significantly reduce energy consumption in actual tasks, it still misses some better options in the experiments of Section \ref{sec5_perf}.
This is because the decision algorithm is based on actual discrete flight data that cannot encompass all situations, causing the system to only approximate the optimal choice.
We can improve the decision using two methods: (1) collecting more data to cover as many situations as possible, and (2) further studying the modeling based on non-discrete method.

\textit{Insight 3:} Although object detection algorithms face numerous challenges in some UAV-based datasets, the detection performance can satisfy actual detection requirements when appropriately trading off other conditions, such as flight altitude, flight speed, and sampling rate.
By carefully designing flight parameters, we can fully utilize object detection techniques to meet various UAV-based object detection requirements.

\textit{Insight 4:} The detection performance of algorithms is inadequate at the edges of images.
Most undetected objects are located at the image edges because they are too small to detect.
We can consider reasonably planning the flight path to position all objects in the detection area within an easily detectable range.
This can improve the detection performance in detection tasks.

\section{Conclusion}
In this study, we propose the E$^3$-UAV system as a means of reducing energy consumption in actual UAV-based object detection tasks.
The system is designed to be compatible with various UAVs, edges, and detection algorithms.
Notably, we have formalized both a task completion score metric and an energy consumption model specifically for real-world tasks.
In addition, we have implemented a lightweight energy efficiency-oriented decision algorithm to assist in selecting appropriate flight parameters.
Experimental results demonstrate that the E$^3$-UAV is capable of effectively performing detection tasks with high energy efficiency, surpassing similar systems and achieving up to an $11.7\%$ reduction in edge energy consumption while simultaneously improving detection performance.
Furthermore, we conclude by offering four insights aimed at furthering the understanding of UAV-based object detection for researchers, with particular attention paid to the significant impact of flight paths on detection performance and energy consumption.
Future work will be devoted to investigating the influence of flight paths on the UAV-based object detection system in order to further enhance its overall performance and efficiency.

\bibliographystyle{IEEEtran}
\bibliography{IEEEabrv, reference}

\vspace{-8 mm}

\begin{IEEEbiography}
[{\includegraphics[width=1in,height=1.25in,clip,keepaspectratio]{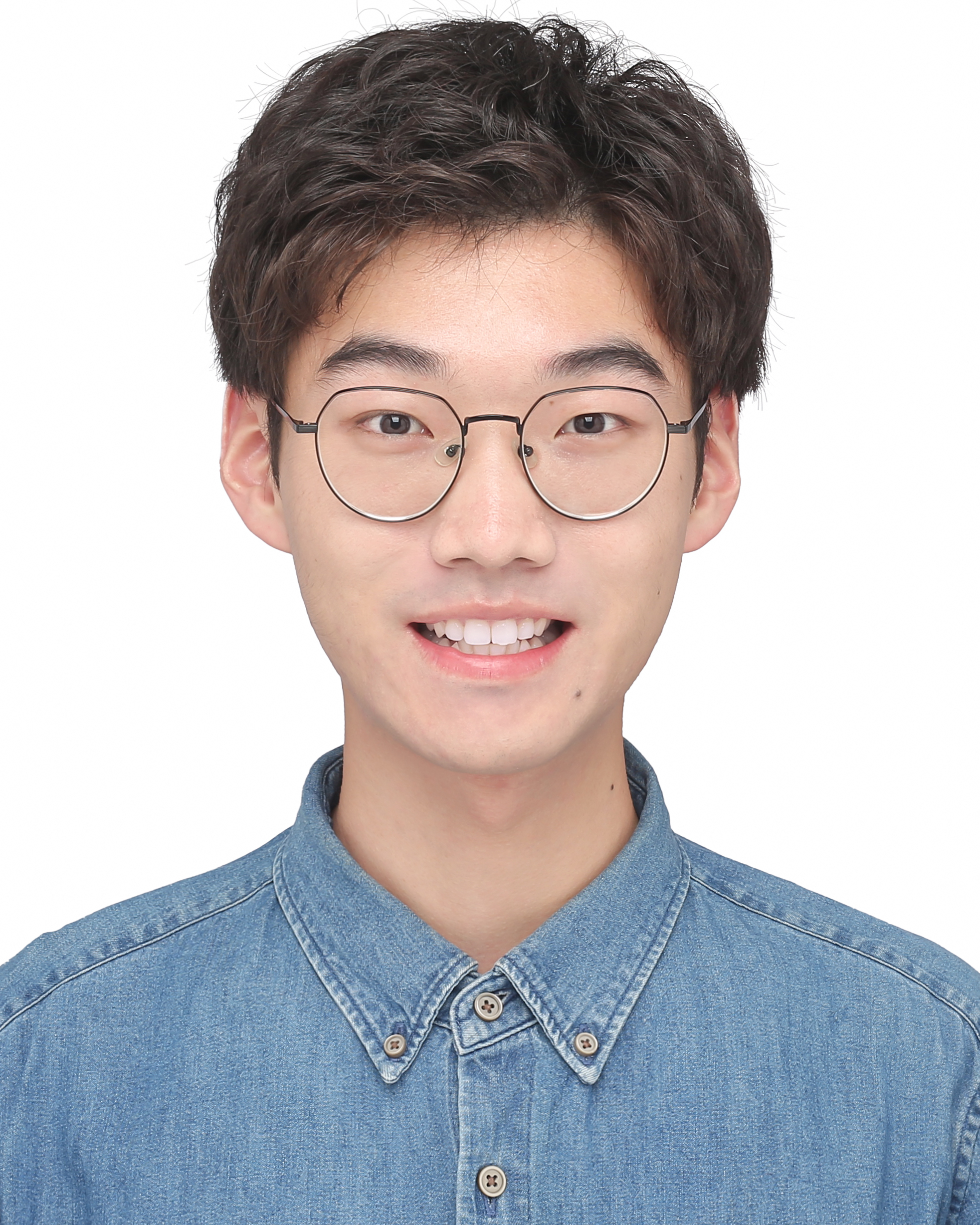}}]{Jiashun Suo}
received the B.Eng. degree in electronic information engineering from Hubei Polytechnic University in 2019, and the M.Eng. degree in software engineering from Yunnan University in 2022. His current research interests include edge computing and machine learning system.
\end{IEEEbiography}

\vspace{-8 mm}

\begin{IEEEbiography}[{\includegraphics[width=1in,height=1.25in,clip,keepaspectratio]{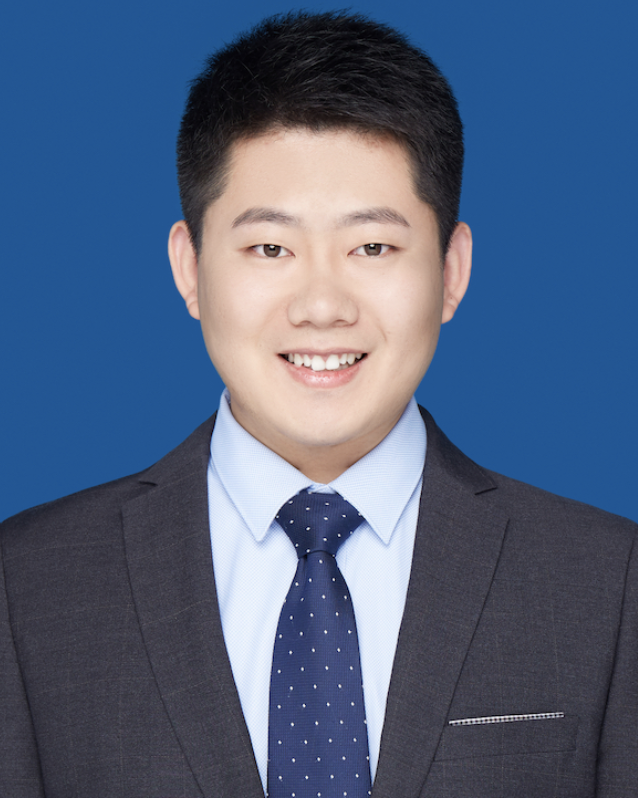}}]{Xingzhou Zhang}
received his Ph.D. degree from Institute of Computing Technology, University of Chinese Academy of Sciences, Beijing. He is an assistant professor of the Institute of Computing Technology, Chinese Academy of Sciences, Beijing. His current research interests include distributed computing systems and edge computing.
\end{IEEEbiography}

\vspace{-8 mm}

\begin{IEEEbiography}[{\includegraphics[width=1in,height=1.25in,clip,keepaspectratio]{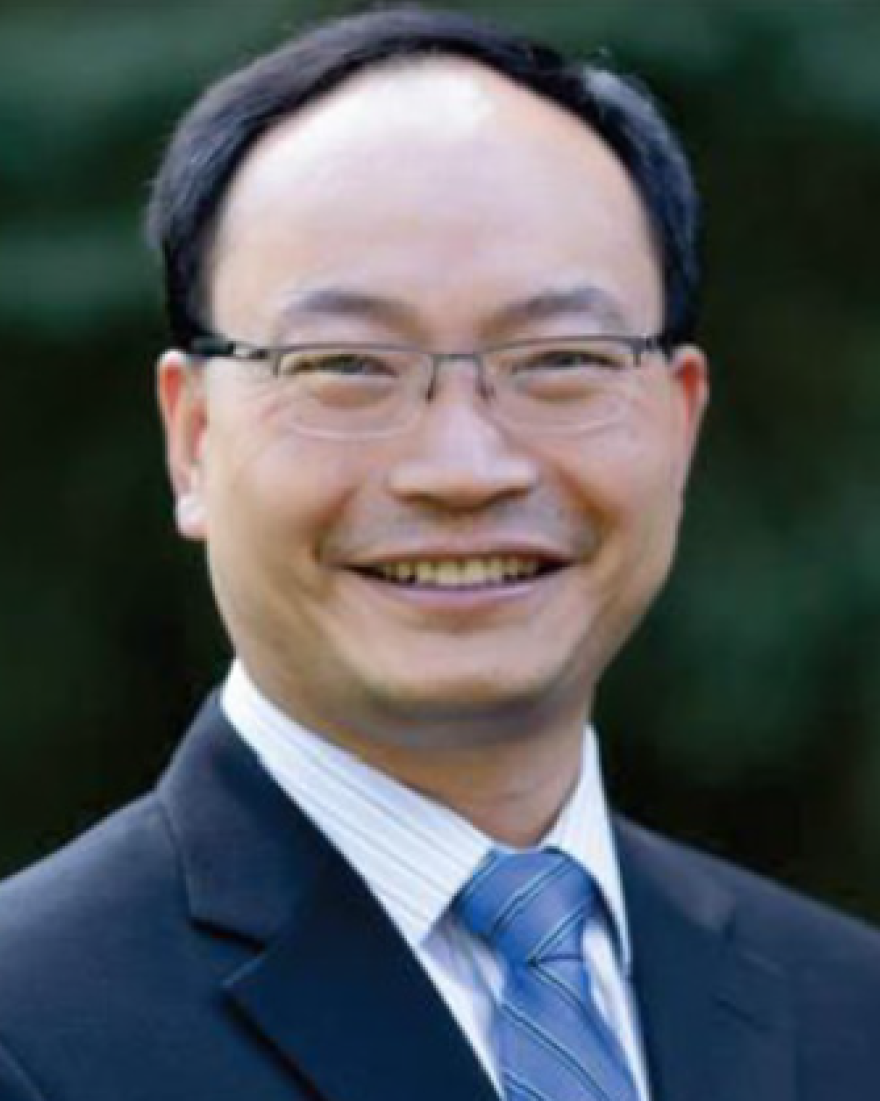}}]{Weisong Shi}
(Fellow, IEEE) is a Professor and the Chair of the Department of Computer and Information Sciences, University of Delaware (UD), where he leads the Connected and Autonomous Research (CAR) Laboratory. He is an internationally renowned expert in edge computing, autonomous driving, and connected health. His pioneer paper titled “Edge Computing: Vision and Challenges” has been cited more than 6200 times. Before he joins UD, he was a Professor with Wayne State University from 2002 to 2022 and served in multiple administrative roles, including an Associate Dean for Research and Graduate Studies with the College of Engineering and an Interim Chair of the Computer Science Department. He also served as a National Science Foundation (NSF) Program Director from 2013 to 2015 and the Chair of two technical committees of the Institute of Electrical and Electronics Engineers (IEEE) Computer Society. He has published more than 270 articles in peer-reviewed journals and conferences and served in editorial roles for more than ten academic journals and publications, including EIC of \textit{Smart Health}, AEIC of IEEE \textit{Internet Computing Magazine}. He is a Distinguished Member of ACM. He is the Founding Steering Committee Chair of several conferences, including ACM/IEEE Symposium on Edge Computing (SEC), IEEE/ACM International Conference on Connected Health (CHASE), and IEEE International Conference on Mobility (MOST). More information is available at http://weisongshi.org.
\end{IEEEbiography}

\vspace{-8 mm}

\begin{IEEEbiography}[{\includegraphics[width=1in,height=1.25in,clip,keepaspectratio]{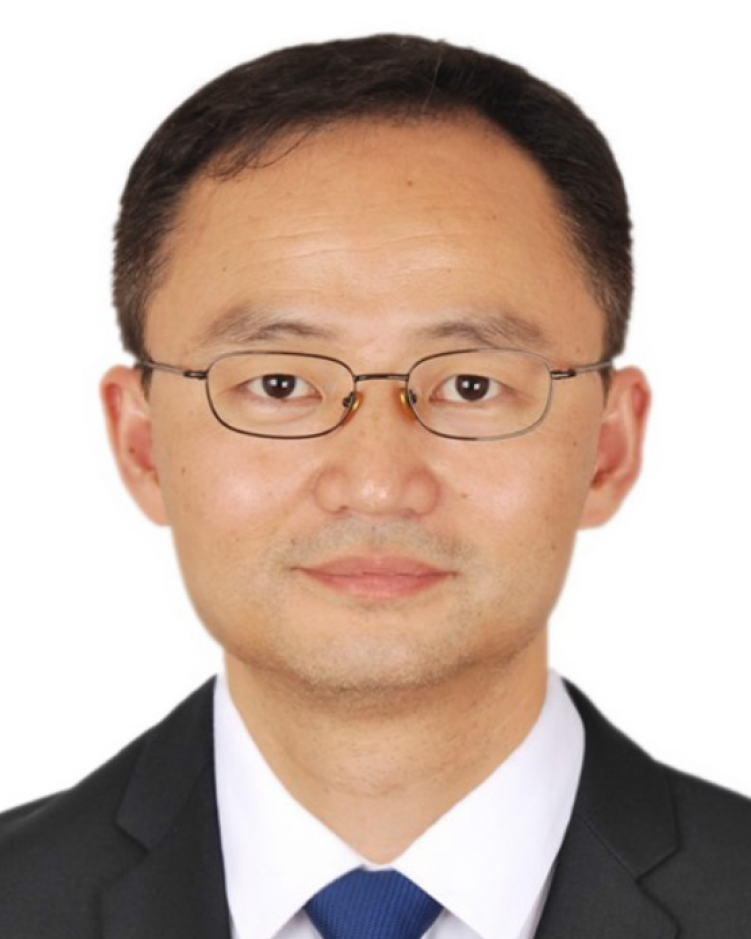}}]{Wei Zhou}
received the Ph.D. degree from the University of Chinese Academy of Sciences. He is currently a Full Professor with the Software School, Yunnan University. His current research interests include the distributed data intensive computing and bioinformatics. He is currently a Fellow of the China Communications Society, a member of the Yunnan Communications Institute, and a member of the Bioinformatics Group of the Chinese Computer Society. He won the Wu Daguan Outstanding Teacher Award of Yunnan University in 2016, and was selected into the Youth Talent Program of Yunnan University in 2017. Hosted a number of National Natural Science Foundation projects.
\end{IEEEbiography}

\end{document}